\documentclass[10pt,twocolumn,letterpaper]{article}

\usepackage{wacv}
\usepackage{customization}

\usepackage[toc,page]{appendix}
\newcommand{\stoptocwriting}{%
  \addtocontents{toc}{\protect\setcounter{tocdepth}{-5}}}
\newcommand{\resumetocwriting}{%
  \addtocontents{toc}{\protect\setcounter{tocdepth}{\arabic{tocdepth}}}}

\def\wacvPaperID{4} % Enter the WACV Paper ID here

\wacvfinalcopy % *** Uncomment this line for the final submission

\ifwacvfinal
\fi

\ifwacvfinal
\usepackage[colorlinks,breaklinks=true,bookmarks=false]{hyperref}
\else
\usepackage[pagebackref=true,breaklinks=true,colorlinks,bookmarks=false]{hyperref}
\fi

\hypersetup{
    linkcolor=eq-fig-tab-color,  % color of internal links 
    citecolor=ref-color,  % color of links to bibliography
}
\usepackage[sort&compress,numbers]{natbib}

\begin{document}

%%%%%%%%% TITLE
\title{Cross-Domain Video Anomaly Detection without Target Domain Adaptation}

\author{Abhishek Aich$^\star$, Kuan-Chuan Peng$^\dagger$, Amit K. Roy-Chowdhury$^\star$\\
$^\star$University of California, Riverside, USA, $^\dagger$Mitsubishi Electric Research Laboratories, USA \\
\small{\texttt{\{aaich001@, amitrc@ece.\}ucr.edu}, 
\texttt{kpeng@merl.com}}
}

\maketitle
\thispagestyle{empty}

%%%%%%%%% ABSTRACT
\begin{abstract}
    \noindent
    Most cross-domain unsupervised Video Anomaly Detection (VAD) works assume that at least few task-relevant target domain training data are available for adaptation from the source to the target domain. However, this requires laborious model-tuning by the end-user who may prefer to have a system that works ``out-of-the-box." To address such practical scenarios, we identify a novel target domain (inference-time) VAD task where no target domain training data are available. To this end, we propose a new `\textit{Zero-shot Cross-domain Video Anomaly Detection} (\zvad)' framework that includes a future-frame prediction generative model setup. Different from prior future-frame prediction models, our model uses a novel \textit{Normalcy Classifier module} to learn the features of normal event videos by learning how such features are different ``relatively" to features in pseudo-abnormal examples. A novel \textit{Untrained Convolutional Neural Network based Anomaly Synthesis} module crafts these pseudo-abnormal examples by adding foreign objects in normal video frames with no extra training cost. With our novel relative normalcy feature learning strategy, \zvad generalizes and learns to distinguish between normal and abnormal frames in a new target domain without adaptation during inference. Through evaluations on common datasets, we show that \zvad outperforms the state-of-the-art (SOTA), regardless of whether task-relevant (\textit{i.e.}, VAD) \textit{source} training data are available or not. Lastly, \zvad also beats the SOTA methods in inference-time efficiency metrics including the model size, total parameters, GPU energy consumption, and GMACs.
\end{abstract}
\captionsetup[figure]{list=no}
\captionsetup[table]{list=no}
\stoptocwriting

%%%%%%%%% BODY TEXT
%----------- INTRODUCTION -----------% 
\section{Introduction}
\begin{figure}[t]
    \centering
    \includegraphics[width=\columnwidth,keepaspectratio]{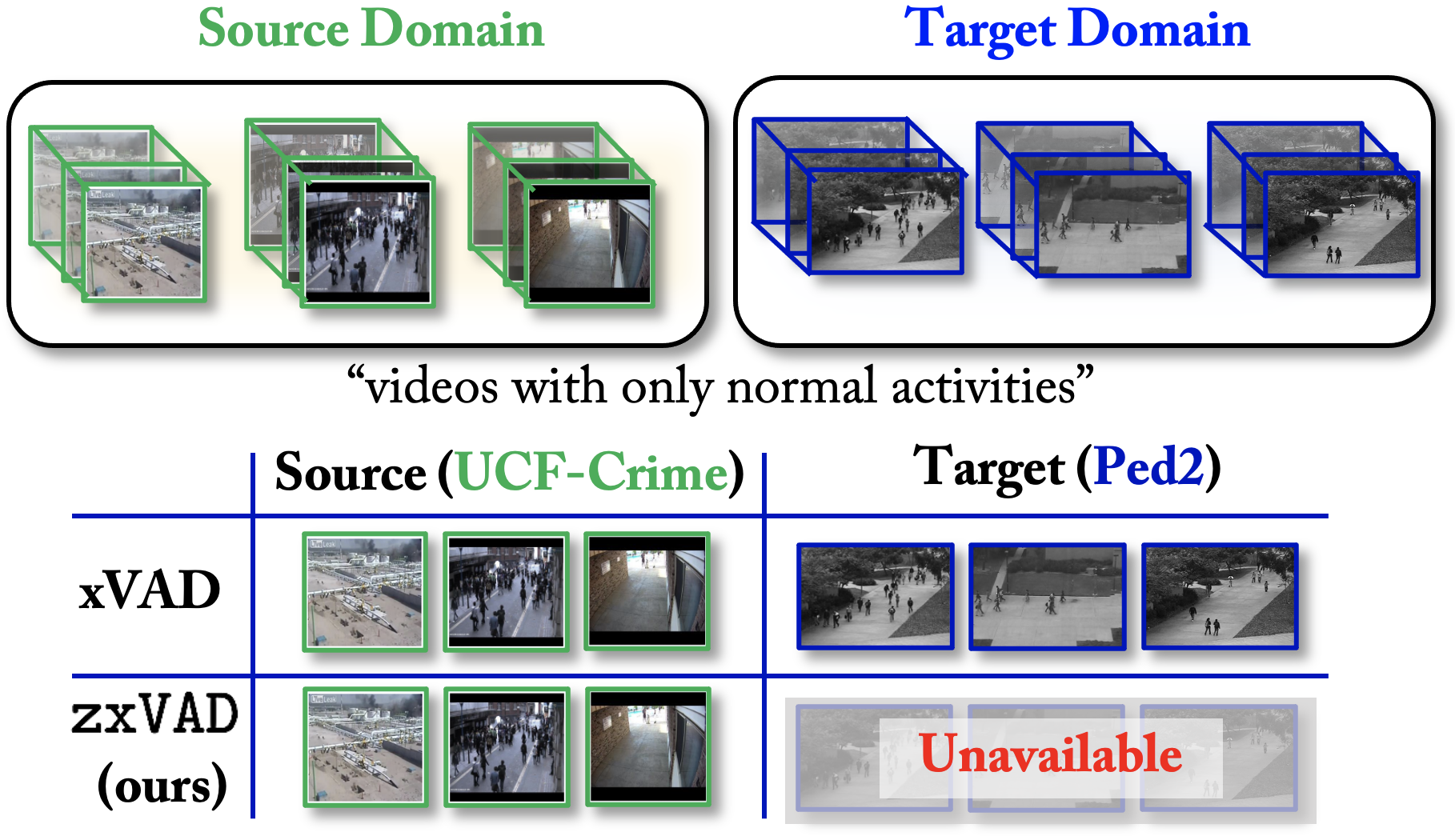}
    \caption{\textbf{Problem overview.} Current unsupervised cross-domain VAD works (xVAD) entail adapting to the target domain, assuming access to at least a few training examples \cite{lu2020few, lv2021learning}. We relax this assumption of having such access to training data from the target domain and tackle a more stringent, yet practical, case using our proposed zero-shot xVAD or \zvad framework.}
    \label{fig:teaser}
    \vspace{-\baselineskip}
\end{figure}

Unsupervised Video Anomaly Detection (VAD) methods \cite{liu2018future, tung2011goal, wu2010chaotic, zhang2005semi, adam2008robust, kim2009observe, lu2013abnormal, mahadevan2010anomaly, cong2011sparse, zhao2011online,xu2017detecting, tudor2017unmasking, nguyen2019anomaly, luo2017revisit, ravanbakhsh2017abnormal, gong2019memorizing, ramachandra2020street, dong2020dual, park2020learning, hasan2016learning, yu2022deep, doshi2022multi, park2022fastano, zaheer2022generative, cho2022unsupervised, lin2022causal, zhang2022adaptive, wang2022unsupervised, shen2022video, ouyang2022look, lee2022multi} have been widely used in security and surveillance applications \cite{pang2021deep, bulusu2020anomalous, chalapathy2019deep, Aich_2021_ICCV} over the supervised or weakly-supervised VAD methods \cite{sultani2018real, feng2021mist, liu2019margin, tian2021weakly, purwanto2021dance, wananomaly, georgescu2021anomaly, park_2020_ACCV, barbalau2022ssmtl++}. This is mainly because unsupervised VAD methods do not need training videos containing abnormal events which are rare and laborious to annotate \cite{bulusu2020anomalous, chalapathy2019deep}. Hence, with only normal events in training videos, the unsupervised VAD methods mark the activities unexplained by the trained model as anomalies during testing. Recently, unsupervised VAD works under cross-domain settings have been introduced \cite{roy2021predicting, georgescu2020background, lu2020few, lv2021learning}. Given the video data containing only normal events from a source domain, the goal is to perform VAD in a different target domain. \tblue{However, these cross-domain VAD (xVAD) works \cite{roy2021predicting, georgescu2020background, lu2020few, lv2021learning} are methods which need access to \underline{either} the source and target domain VAD training data \cite{lu2020few, lv2021learning} or strong supervision from pre-trained object detectors (\eg, YOLOv3 \cite{redmon2018yolov3} in \cite{georgescu2020background})}. {Collecting such data in the target domain and adapting or tuning the model may not be feasible by the end-user who may want a system that works ``out-of-the-box" \cite{moriya2018evolution, liu2019auptimizer}. Moreover, granting access to such video data may be time-consuming to third-party corporations due to intellectual property and security concerns \cite{krotova2020open, mattioli2014disclosing}.} This renders the current xVAD works ineffective as they assume access to at least some target domain training data.
\paragraph{Problem Statement.} Based on the aforesaid issues, we formally identify the following new unsupervised xVAD problem of detecting anomalies in the target domain with strictly \emph{no access to target domain training data and no prior knowledge of its anomaly types}. More specifically, our goal is to detect anomalies in the \textit{target} domain's testing set, without having any training data on the target side. Fig.~\ref{fig:teaser} contrasts this problem setup with prior xVAD problem definitions.
\paragraph{Proposed framework.} We tackle this new problem using a novel xVAD framework, namely `Zero-shot Cross-domain Video Anomaly Detection' (\zvad). The term \textit{zero-shot} implies \textit{no training videos available} from the target domain for adaptation to perform anomaly detection. {\zvad has a generator \cite{goodfellow2020generative, Aich_2020_CVPR, gupta2020alanet} in a future-frame prediction setup \cite{liu2018future} similar to xVAD approaches \cite{lv2021learning, lu2020few}. However different from these methods, \zvad's generator training is assisted by a novel \textit{Normalcy Classifier} (NC) module and an \textit{Untrained Convolutional Neural Network (CNN) Anomaly Synthesis} ($\bm{\mathcal{O}}$) module. Prior unsupervised xVAD works learn features from \textit{only} videos with normal events. This leads to overfitting to the source domain distribution and the poor generalizing ability for target domain VAD \cite{lv2021learning}. In contrast, \zvad's generator uses NC and $\bm{\mathcal{O}}$ modules to learn features of normal activities in input videos, by focusing on how such features are \textit{relatively} different from features of abnormal frames. This ``relative" learning strategy enhances the generator's ability in identifying anomalies in target domain without any adaptation at test time.

$~~~$Normalcy, by definition, is always contextually dependent \cite{wananomaly, ionescu2019object} (\eg, running in playgrounds \textit{vs} highways). Hence, to generalize across new target domains (without its training data) where we have no prior knowledge about anomaly types, we propose to learn normal event features that consider the contextual or relative difference between ``normal" and ``abnormal" patterns. {More concretely, rather than learning only the normalcy features (\ie features of normal video frames), our model learns the \textit{relative} normalcy features (\ie difference between features of normal \textit{and} abnormal video frames) using our proposed NC module.} These pseudo-abnormal frames are created through our proposed
$\bm{\mathcal{O}}$ module which is capable of localizing objects from both Task-Relevant or VAD data and Task-Irrelevant (TI) or non-VAD data (\ie, data irrelevant to the VAD task). The $\bm{\mathcal{O}}$ module crafts pseudo-abnormal frames by localizing objects in input TI or VAD video frames and pasting them (with random location and size) on normal VAD video frames. Furthermore, a major advantage of introducing TI data to our problem setup is that they can be treated as \textit{video distributions for learning patterns of normal activities} and also assist in creating diverse anomalies. Hence, along with the strategy of learning the relative normalcy difference, \zvad aims to mitigate the generalizing issue \textit{via} learning this relative normalcy with respect to abnormal frames having different kinds of foreign objects (either from VAD or TI frames). This allows \zvad to avoid being limited to specific anomaly types in the source domain, making it fundamentally different from supervised specific anomaly learning. 

Our NC module is designed to distinguish between a pseudo-abnormal and the predicted normal future-frame through novel loss functions. The highlighting attribute of these functions is to consider different properties of normal and abnormal frames through our NC's logit predictions and derived attention maps. {Our $\bm{\mathcal{O}}$ module is uniquely capable of using VAD or TI data with an \ul{untrained randomly initialized} CNN to create anomalies at no extra training cost.}} To sum up, we make the following key \textbf{contributions}:
\begin{enumerate}[label=\arabic*., leftmargin=*, topsep=0pt]
\setlength\itemsep{-0.0em}
    \item We formally introduce a new problem setup in xVAD where the model is trained on the source domain to detect anomalies (at test time) in \textit{a different target domain without any adaptation via target-domain training data} or \tblue{using any supervision from pre-trained models (\eg YOLOv3)}.
    \item A novel xVAD method namely \zvad is proposed, where the model learns the \textit{relative difference between normal and abnormal frames} in source domain and generalize VAD to target without needing target domain training data \tblue{or any external support from pre-trained models.}
    \item This ``\textit{relative}" difference learning is achieved via a novel Normalcy Classifier that uses a new pseudo-anomaly synthesis module based on an untrained CNN where anomalies are created with no extra training cost.
    \item Notably, for the first time in VAD literature (to the best of our knowledge), we also show that \zvad outperforms the SOTA xVAD works in the proposed problem setup when trained \underline{only} with TI data, in four common benchmarks.
    \item \zvad beats the SOTA xVAD works in the proposed problem setup both in AUC on most benchmarks, and in inference-time efficiency metrics (\eg, model size, model parameters, GPU energy consumption, and GMACs).
\end{enumerate}

%----------- RELATED WORKS ----------% 

\begin{table}[!t]
\setlength{\tabcolsep}{10pt}
\caption{\textbf{Characteristic comparison.} Better than prior unsupervised VAD works (\eg, $\mathcal{C}_0$: \cite{hinami2017joint, sun2020scene, yu2020cloze, liu2021hybrid, chen2014crowd, zaheer2020old, ionescu2019object}, $\mathcal{C}_1$: \cite{liu2018future, tung2011goal, wu2010chaotic, zhang2005semi, adam2008robust, kim2009observe, lu2013abnormal, mahadevan2010anomaly, cong2011sparse, zhao2011online,xu2017detecting, tudor2017unmasking, nguyen2019anomaly, luo2017revisit, ravanbakhsh2017abnormal, gong2019memorizing, ramachandra2020street, dong2020dual, park2020learning, hasan2016learning, yu2022deep, doshi2022multi, park2022fastano}, $\mathcal{C}_2$: \cite{roy2021predicting, georgescu2020background, doshi2020any}, $\mathcal{C}_3$ (our baselines): \cite{lu2020few, lv2021learning}), \zvad needs no {prior knowledge (\eg. object extraction from VAD videos)}, can perform cross-domain VAD with no VAD training data, and uses an untrained CNN to create anomalies.}
%\vspace{-.75em}
\resizebox{\columnwidth}{!}{%
\begin{tabular}{lcccc>{\columncolor{red!10}}c}
\toprule
\multirow{2}{*}{\textbf{{Unsupervised VAD Method} Conditions}} & \multicolumn{5}{c}{\textbf{Unsupervised VAD Categories}} \\ \cmidrule(l){2-6} 
 & $\mathcal{C}_0$ & $\mathcal{C}_1$ & $\mathcal{C}_2$ & $\mathcal{C}_3$ & \zvado \\
\midrule
\textbf{no} prior {\textbf{knowledge}} required?  & \ccross & $\ccheck$ & \ccross & $\ccheck$ & $\ccheck$ \\
show efficacy in \textbf{cross-domain} VAD?  & \ccross & \ccross & $\ccheck$ &  $\ccheck$ & $\ccheck$ \\
works with \textbf{no {source domain} VAD training data}? & \ccross & \ccross & \ccross & \ccross & $\ccheck$ \\
create pseudo-anomalies with an \textbf{untrained} network? & \ccross & \ccross & \ccross & \ccross & $\ccheck$ \\
\bottomrule
\end{tabular}%
}
%\vspace{-.5em}
\label{tab:related_works}
\vspace*{-\baselineskip}
\end{table} 
\section{Related Works}
\vspace{0.8em}
\paragraph{Unsupervised VAD works.} Early unsupervised VAD works formulated the anomaly detection using handcrafted features to characterize the normal event or regular pattern distribution \cite{tung2011goal, wu2010chaotic, zhang2005semi, adam2008robust, kim2009observe, lu2013abnormal, mahadevan2010anomaly, cong2011sparse, zhao2011online}. However, these methods were outperformed by the CNN approaches \cite{park2020learning, xu2017detecting, dong2020dual, tudor2017unmasking, nguyen2019anomaly, hasan2016learning, luo2017revisit, ravanbakhsh2017abnormal, gong2019memorizing, liu2018future, ramachandra2020street, yu2022deep, doshi2022multi, park2022fastano} (both categorized as $\mathcal{C}_1$ in Tab. \ref{tab:related_works}). Some of these CNN based unsupervised VAD works use generators \cite{goodfellow2020generative} to model the normal frame distributions \cite{liu2018future, dong2020dual, park2020learning, hasan2016learning, gong2019memorizing, liu2021hybrid}, and further introduce memory modeling networks to record various normal event patterns in videos \cite{lu2020few, liu2021hybrid, lv2021learning, park2020learning, gong2019memorizing}. Another category of works ($\mathcal{C}_0$ in Tab. \ref{tab:related_works}) \cite{hinami2017joint, ionescu2019object, sun2020scene, yu2020cloze, liu2021hybrid, chen2014crowd, zaheer2020old} proposed computationally heavy approaches that used strong priors like object extraction (using pre-trained object-detectors \cite{ionescu2019object,yu2020cloze}) for VAD, in order to focus only on specific objects to detect anomalies. Compared to aforesaid VAD works in $\mathcal{C}_0$ and $\mathcal{C}_1$, \zvad \textit{(a)} is designed to tackle unsupervised \textit{cross-domain} VAD problem, \textit{(b)} is a future-frame prediction method with a memory module, and \textit{(c)} needs no strong prior knowledge from object extraction. {Finally, few works \cite{astrid2021synthetic, astrid2021learning, ionescu2019object, georgescu2020background, zaheer2020old, pourreza2021g2d} have shown different VAD strategies where pseudo-anomalies are used. For example, \cite{ionescu2019object, pourreza2021g2d} uses a generator to create fake anomaly data. \cite{astrid2021synthetic, astrid2021learning} propose two different temporal pseudo anomaly synthesizers to craft anomalies from normal videos. Needing no aforesaid extra training efforts, \zvad uses a novel strategy to create anomalies using an \textit{untrained randomly initialized} CNN (details in Sec. \ref{sec:method}) at no extra training cost.}
%\vspace{-1em}
\paragraph{Cross-domain setting.} The cross-domain scenario in unsupervised VAD has been introduced in \cite{roy2021predicting, doshi2020any, lv2021learning, lu2020few, hu2021adaptive}. These works operate under the regime of few-shot target domain scene adaptation. For example, \cite{lu2020few, lv2021learning} ($\mathcal{C}_3$ in Tab. \ref{tab:related_works}) use meta-learning approaches \cite{hospedales2020meta} and adapt to the target domain with few scenes for anomaly detection. In contrast, \zvad is specifically designed for cross-domain VAD \textit{without any target domain adaptation}. \cite{roy2021predicting, georgescu2020background, doshi2020any}  ($\mathcal{C}_2$ in Tab. \ref{tab:related_works}) provide prior knowledge based methods where videos are subject to object-extraction using pre-trained object detectors \cite{redmon2018yolov3, lin2017feature}. However, \zvad needs no strong priors like object extraction using pre-trained detectors. \zvad is also capable of solely using TI data and outperforming the SOTA in proposed cross-domain VAD setup. Finally, \zvad uses a simple training strategy (details in Sec. \ref{sec:method}) rather than using a meta-learning approach to avoid non-trivial computational and memory burdens, as well as vanishing gradients issues \cite{rajeswaran2019meta, roy2021predicting}.

%---------- PROPOSED WORK -----------% 
\section{Proposed {\normalfont{\zvad}} Framework}\label{sec:method}
\begin{figure*}[t]
    \vspace*{-1\baselineskip}
    \centering
    \includegraphics[width=\textwidth]{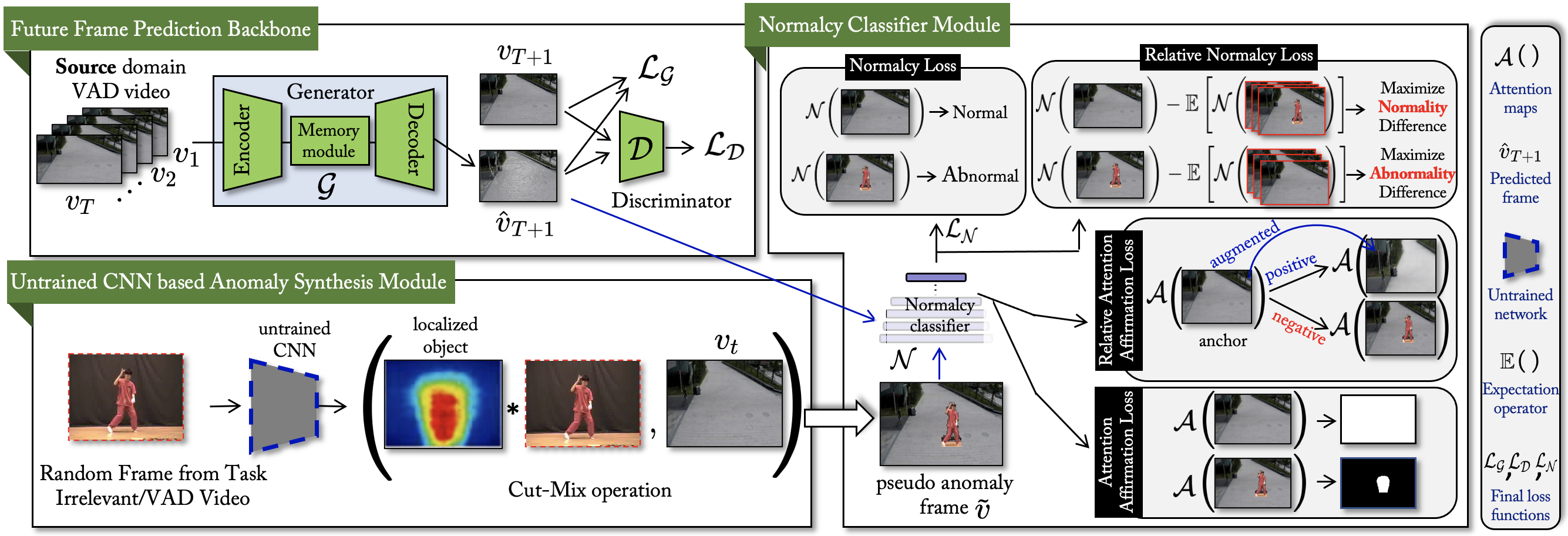}
    \caption{\textbf{Framework overview.} Our \zvad framework contains a Future-Frame Prediction backbone (\textit{top-left}) guided by our Normalcy Classifier module (\textit{right}). To enforce the prediction backbone to learn generalizable features from source domain normal videos and avoid overfitting, we encourage the generative model to learn normalcy features relative to pseudo-abnormal frames using four novel loss functions. These abnormal frames are created using an untrained randomly-initialized CNN through our novel anomaly synthesis module $\modO$ (\textit{bottom-left}).}
    \label{fig:main_framework}
    \vspace*{-1\baselineskip}
\end{figure*}

\vspace{0.8em}
\paragraph{Method Overview.} We are provided with source domain VAD normal videos to learn features that should ideally transfer across different target domains without needing target domain adaptation. To achieve this \textit{no} adaptation-based cross-domain VAD property, we introduce a novel \zvad framework (illustrated in Fig. \ref{fig:main_framework}) based on a future-frame prediction setup that can be trained end-to-end. It consists of an untrained CNN based pseudo-anomaly Synthesis module (Sec. \ref{sec:pseudo_anomaly}) where an untrained randomly initialized CNN helps in creating pseudo-anomalies without any extra training burden. These pseudo-abnormal frames along with predicted future-frame are utilized in our novel Normalcy Classifier module (Sec. \ref{sec:normalcy_classifier}) to regularize the backbone generator to learn relative normalcy features. This learning strategy makes \zvad capable of more generalizable VAD performance across different target domains than existing xVAD methods.

\paragraph{Notations.} We denote a sample video from VAD datasets as $[\bm{v}_1, \bm{v}_2, \cdots, \bm{v}_{L_v}]\in\mathbb{R}^{L_v\times C\times H\times W}$, and TI datasets as $[\bm{u}_1, \bm{u}_2, \cdots, \bm{u}_{L_u}]\in\mathbb{R}^{L_u\times C\times H\times W}$, where each video contains $L_v$ and $L_u$ number of frames, and each frame is of height $H$, width $W$, and $C$ channels. Our future-frame prediction framework \zvad contains a memory-augmented generator \cite{gong2019memorizing} $\mathcal{G}(\cdot)$ with weights $\thetaG$ and memory module $\mathcal{M}$, and a discriminator $\mathcal{D}(\cdot)$ with weights $\thetaD$. As shown in \cite{gong2019memorizing, weston2014memory, rae2016scaling}, the memory module $\mathcal{M}\in\mathbb{R}^{K\times Q}$ is a matrix with $\bm{m}_i\in\mathbb{R}^Q,\forall i\in[K]$ vectors (or memory items) that learns to register the prototypical normal features during training. $\mathcal{M}$ takes the output vector $\bm{z}\in\mathbb{R}^Q$ from $\mathcal{G}(\cdot)$'s encoder and outputs $\widehat{\bm{z}}=\bm{w}\mathcal{M}\in\mathbb{R}^Q$ that is forwarded to $\mathcal{G}(\cdot)$'s decoder. Here, $\bm{w}\in\mathbb{R}^{1\times K}$ is termed as a soft addressing vector \cite{gong2019memorizing}. Each element $w_i$ of $\bm{w}$ is computed using softmax operation on the cosine similarity between $\bm{z}$ and $\bm{m}_i$ \cite{gong2019memorizing, santoro2016one}. Our proposed anomaly synthesis module is denoted as $\bm{\mathcal{O}}$ and contains a CNN denoted as $\mathcal{R}(\cdot)$ with weights $\thetaR$. Further, our proposed normalcy classifier module contains a CNN  classifier denoted as $\mathcal{N}(\cdot)$ with weights $\thetaN$. We denote the expectation operator, $l_p$-norm operator, and element-wise multiplication by $\mathbb{E}[\cdot]$, $\Vert\cdot\Vert_p$, and $\odot$, respectively.
\paragraph{Backbone description.} Given $N$ source domain training videos (with only normal events), we aim to learn a future-frame prediction generator that takes in $T$ input frames and predicts a future frame $\widehat{\bm{v}}_{T+1}$, \ie, $\mathcal{G}\big{(}[\bm{v}_1, \bm{v}_2, \cdots, \bm{v}_{T}]\big{)}$ = $\widehat{\bm{v}}_{T+1}$. $\mathcal{G}(\cdot)$ is adversarially trained against $\mathcal{D}(\cdot)$ in the Least-Square GAN \cite{mao2017least, goodfellow2020generative} setup where $\mathcal{D}(\cdot)$ aims to distinguish between $\widehat{\bm{v}}_{T+1}$ and the ground truth frame $\bm{v}_{T+1}$. Similar to \cite{gong2019memorizing, lv2021learning}, we introduce a memory module $\mathcal{M}$. In \zvad, $\mathcal{G}(\cdot)$ is further regularized using $\mathcal{N}(\cdot)$ with our proposed four novel objectives (explained in Sec. \ref{sec:normalcy_classifier}) which uses pseudo-anomaly examples generated using an untrained CNN based strategy. Following prior works \cite{mathieu2015deep, lu2020few}, we optimize $\mathcal{G}(\cdot)$ with the mean square error loss $\mathcal{L}_{\text{MSE}} = \Vert\widehat{\bm{v}}_{T+1} - \bm{v}_{T+1} \Vert^2_2$, structure similarity loss $\mathcal{L}_{\text{SSM}} = 1 - \text{SSIM}\big{(}\widehat{\bm{v}}_{T+1}, \bm{v}_{T+1}\big{)}$, where SSIM represents the structural similarity index measure \cite{wang2004image} between $\widehat{\bm{v}}_{T+1}$ and $\bm{v}_{T+1}$, and Gradient loss $\mathcal{L}_{\text{GD}}$ \cite{mathieu2015deep, liu2018future}. To optimize $\mathcal{M}$ and encourage modeling normal videos using sparse but most relevant memory slots, we follow \cite{gong2019memorizing} and apply a hard-shrinkage on $\mathcal{M}$'s memory addressing vectors $w_i$ using continuous ReLU activation function with a shrinkage factor $\lambda$ set as 0.0005. Next, we normalize each element $\widehat{w}_i \leftarrow \nicefrac{\widehat{w}_i}{\Vert\bm{\widehat{w}}\Vert_1}\forall i$ and get $\widehat{\bm{z}}=\widehat{\bm{w}}\mathcal{M}$. We also apply a sparsity regularizer on $\widehat{\bm{w}}$ by minimizing its entropy as $\mathcal{L}_{\text{MEM}} = \sum_{i=1}^N -\widehat{w}_i\log\big{(}\widehat{w}_i\big{)}$ \cite{gong2019memorizing}. We combine these losses as 
\begin{align}
    \mathcal{L}_{\text{BB}} = \mathcal{L}_\text{REC} + \alpha_\text{MEM}\mathcal{L}_\text{MEM},
\end{align}
where the reconstruction loss is $\mathcal{L}_\text{REC} = \mathcal{L}_\text{MSE} + \mathcal{L}_\text{SSM} + \mathcal{L}_\text{GD}$. We set the loss weight $\alpha_\text{MEM}=0.0025$ following \cite{gong2019memorizing}. Totally, the weights $\thetaG$, $\thetaD$ and $\thetaN$ are updated during training, while $\thetaR$ is randomly initialized before training and remains fixed. Better than prior works which do not consider the relative difference between normal and abnormal events, \zvad introduces a novel strategy to regularize this backbone generator by learning normal features with respect to pseudo-abnormal features. As our normalcy classifier module utilizes pseudo-anomalies to learn the \textit{relative} normalcy features, we first present our pseudo-anomaly creation strategy. 

\subsection{Pseudo-Anomaly Synthesis via Untrained CNN} \label{sec:pseudo_anomaly}
Prior works \cite{ionescu2019object} have focused on creating anomalies using pre-trained object detectors (\ie, YOLOv3 \cite{redmon2018yolov3} in \cite{ionescu2019object}) that result in issues like additional training overheads. Different from such methods, we present a training-free strategy to extract objects from video frames. These objects can be obtained on both VAD and TI video frames (\ie, $\bm{v}_t$ and $\bm{u}_t$). For brevity, we refer to the input frame as $\bm{x}$. Given an input frame $\bm{x}\in\mathbb{R}^{C\times H\times W}$, we denote the output of a CNN $\mathcal{R}(\cdot)$ (before the classification layer) as tensor $\bm{G}\in\mathbb{R}^{d\times h\times w}$. For example, if $\mathcal{R}(\cdot)$ is ResNet152 \cite{he2016deep}, $\bm{G}$ is the output of `$conv5\_x$' with size ${2048\times 8\times 8}$ if input size is ${3\times 256\times 256}$. We employ SCDA \cite{wei2017selective} to perform channel-wise summation on $\bm{G}$ to obtain an attention map $\bm{A}\in\mathbb{R}^{h\times w}$. We then obtain a binary mask $\bm{M}$ from $\bm{A}$ as follows. We set $\bm{M}_{(i,j)} = 1 \text{ if }\bm{A}_{(i,j)}>\varsigma$, or 0 otherwise. Here, $(i,j)$ represents position in $h\times w$ locations. We empirically set $\varsigma = 0.1$. $\bm{M}_{(i,j)}=1$ indicates the foreground objects. Finally, $\bm{M}$ is resized from $h\times w$ to $H\times W$. As noted in \cite{cao2021random}, the idea behind this surprising property that randomly initialized CNN can localize objects is: because the background in the input frame $\bm{x}$ is relatively texture-less in comparison to the foreground objects in the scene, these background regions have higher chances to be deactivated by nonlinear activation functions like ReLU \cite{hahnloser2000digital}. The object is finally localized as $\bm{M}_{\bm{x}}=\bm{M}\odot\bm{x}$. 
%\paragraph{Synthesizing anomalies.}
To create pseudo-abnormal frame $\tilde{\bm{v}}$, we combine $\bm{M}_{\bm{x}}$ and one of the input frames to $\mathcal{G}(\cdot)$, \ie, $\bm{v}_t \in \left \{ \bm{v}_1, \bm{v}_2, \cdots , \bm{v}_{T} \right \}$ by pasting $\bm{M}_{\bm{x}}$ on $\bm{v}_t$ at random location $r_z$ with random size $r_x\times r_y$. We discuss the method to choose the location $r_z$ and size $r_x\times r_y$ in Supplementary Material. Note that most of the video frames used for creating pseudo-anomalies happen to contain at least one foreground object for the untrained CNN to extract. Even if there are no such objects, our untrained CNN will still focus on some patches (on the input frame) and treat them as anomaly on normal event VAD frame.
\begin{figure}[t]
    \vspace*{-0.75\baselineskip}
    \centering
    \includegraphics[width=1\columnwidth]{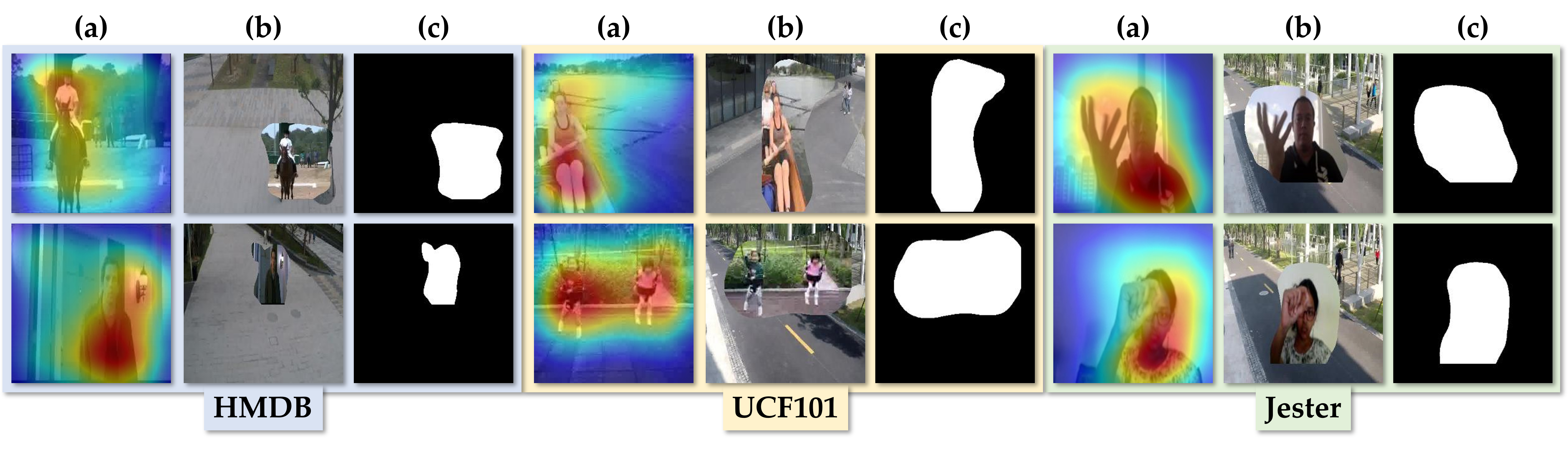}
    %\vspace{-2.5em}
    \caption{\textbf{Pseudo-abnormal examples.} We show pseudo-abnormal frames (marked as \textbf{(b)}) created using our pseudo-anomaly synthesis strategy. The untrained randomly initialized CNN is ResNet50 \cite{he2016deep} which localizes objects in the TI frames (marked as \textbf{(a)}). We also show examples of ground-truth masks $\tilde{\bm{M}}$ used in loss $\mathcal{L}_{\text{RAA}}$ (marked as \textbf{(c)}). See more examples in Supplementary Material.}
    \label{fig:pseudo_anomaly}
    \vspace*{-1.25\baselineskip}
\end{figure}
\subsection{Learning Normality w.r.t. Abnormality}\label{sec:normalcy_classifier}
\noindent Our Normalcy Classifier Module is a classifier $\mathcal{N}(\cdot)$ that is optimized by the following four loss functions. These loss functions are complementary to each other as follows: \textit{normalcy} loss and \textit{attention affirmation} loss focus on the difference between normal and abnormal frames, whereas \textit{relative normalcy} loss and \textit{relative attention affirmation} loss focus on how relatively different are normal frames from abnormal frames (and vice-versa). For clarity, we drop the subscript of the predicted frame $\widehat{\bm{v}}_{T+1}$ and mark it as $\widehat{\bm{v}}$. The data distribution of normal and pseudo-abnormal frames are denoted as $\rho$ and $\kappa$, respectively.\\
\noindent \textbf{Normalcy loss $\mathcal{L}_{\text{N}}$.} Given the predicted future-frame $\widehat{\bm{v}}$ and pseudo-abnormal frame $\tilde{\bm{v}}$, $\mathcal{L}_{\text{N}}$ optimizes $\mathcal{N}(\cdot)$ to increase the probability that  $\widehat{\bm{v}}$ is `normal' (label set as 1) and $\tilde{\bm{v}}$ is `abnormal' (label set as 0), using following loss function.
\begin{align}
    \small{
    \textstyle{
    \mathcal{L}_{\text{N}} = \nicefrac{1}{2}\mathbb{E}_{\widehat{\bm{v}}\sim \rho}\Big{[}\big{(}\mathcal{N}(\widehat{\bm{v}})-1\big{)}^2\Big{]} + \nicefrac{1}{2}\mathbb{E}_{\tilde{\bm{v}}\sim \kappa}\Big{[}\big{(}\mathcal{N}(\tilde{\bm{v}})\big{)}^2\Big{]}}}
    \label{eq:normlacy_loss}
\end{align}
\noindent \textbf{Relative normalcy loss $\mathcal{L}_{\text{RN}}$.} Abnormal events can be viewed as deviation with respect to normal events. We argue that the key missing attribute of \eqref{eq:normlacy_loss} is that the probability of normal data being \textit{normal} ($\mathcal{N}(\widehat{\bm{v}})$) should increase as the probability of abnormal data being \textit{normal} ($\mathcal{N}(\tilde{\bm{v}})$) decreases and vice-versa. Rather than just maximizing $\mathbb{P}[\widehat{\bm{v}} \text{ is normal}]$, we also ask $\mathcal{N}(\cdot)$ to maximize $\mathbb{P}[\widehat{\bm{v}} \text{ is more normal than } \tilde{\bm{v}}]$ ($\mathbb{P}[\cdot]$ denotes probability operator). We define this novel relative normalcy loss below: 
\begin{align}
{
\begin{gathered}
    \mathcal{L}_{\text{RN}} = \nicefrac{1}{2}\mathbb{E}_{\widehat{\bm{v}}\sim \rho}\Big{[}\big{(}\mathcal{N}(\widehat{\bm{v}})-\mathbb{E}_{\tilde{\bm{v}}\sim \kappa}[\mathcal{N}(\tilde{\bm{v}})]-1\big{)}^2\Big{]} + \\ \nicefrac{1}{2}\mathbb{E}_{\tilde{\bm{v}}\sim \kappa}\Big{[}\big{(}\mathcal{N}(\tilde{\bm{v}})-\mathbb{E}_{\widehat{\bm{v}}\sim \rho}[\mathcal{N}(\widehat{\bm{v}})] + 1\big{)}^2\Big{]}
    \label{eq:rel_normlacy_loss}
\end{gathered}}
\end{align}

\noindent \textbf{Attention affirmation loss $\mathcal{L}_{\text{AA}}$.} The decision of $\mathcal{N}(\cdot)$ on the normal frame $\widehat{\bm{v}}$ and abnormal frame $\tilde{\bm{v}}$ should be based on the following information: \textit{(1)} $\mathcal{N}(\cdot)$ should consider the whole scene in $\widehat{\bm{v}}$ to classify it as `\textit{normal},' and \textit{(2)} $\mathcal{N}(\cdot)$ should consider the foreign object (introduced by our module $\modO$ in $\tilde{\bm{v}}$) to classify it as `\textit{abnormal}.' Our strategy in Sec. \ref{sec:pseudo_anomaly} allows us to obtain the exact location of foreign objects in $\tilde{\bm{v}}$. Hence, we leverage this knowledge and create ground-truth masks of $\tilde{\bm{v}}$. We first initialize a tensor $\tilde{\bm{M}}$ with zeroes. Next, we update this tensor by pasting $\bm{M}$ after resizing to $r_x\times r_y$ at location $r_z$ (obtained from $\modO$ in Sec. \ref{sec:pseudo_anomaly}). We show examples of $\tilde{\bm{M}}$ in Fig. \ref{fig:pseudo_anomaly}. We extract feature maps from the last convolutional layer of $\mathcal{N}(\cdot)$ and apply SCDA \cite{wei2017selective} to obtain attention maps $\mathcal{A}(\widehat{\bm{v}})$ and $\mathcal{A}(\tilde{\bm{v}})$ for normal and abnormal frames, respectively. $\mathcal{A}(\cdot)$ denotes the operation to extract attention maps from $\mathcal{N}(\cdot)$. We enforce this constraint via the attention affirmation loss $\mathcal{L}_{\text{AA}}$ as ($\mathbb{1}$ is a tensor of the same size as $\mathcal{A}(\widehat{\bm{v}})$ filled with ones):
\begin{align}
     \mathcal{L}_{\text{AA}} = \nicefrac{1}{2}\big{(}\mathbb{1} - \mathcal{A}(\widehat{\bm{v}})\big{)}^2 + \nicefrac{1}{2}\big{(}\tilde{\bm{M}} - \mathcal{A}(\tilde{\bm{v}})\big{)}^2,
     \label{eq:att_aff_loss}
\end{align}
\vspace{-\baselineskip}
\paragraph{Relative attention affirmation loss $\mathcal{L}_{\text{RAA}}$.} Similar to the concept of $\mathcal{L}_{\text{RN}}$, we argue that $\mathcal{L}_{\text{AA}}$ does not consider the relative difference of attention maps from normal frames with respect to attention maps from abnormal frames. Hence, we propose a relative attention affirmation loss $\mathcal{L}_{\text{RAA}}$ that aims to learn this difference. We create two attention map pairs: \textit{(Pair-1)} $\mathcal{A}(\widehat{\bm{v}})$ and $\mathcal{A}(g(\widehat{\bm{v}}))$, and \textit{(Pair-2)} $\mathcal{A}(\widehat{\bm{v}})$ and $\mathcal{A}(\tilde{\bm{v}})$. The function $g(\cdot)$ denotes a series of transformations (\textit{Color Jitter}, \textit{Random Affine}, and \textit{Random Perspective}) applied to $\widehat{\bm{v}}$ using the package Kornia \cite{riba2020kornia} (related parameters are provided in Supplementary Material). The relative difference between the attention on `augmented normal' frame should be smaller than that of the `pseudo-abnormal' frame with respect to the `normal' frame. {We enforce this difference with a margin $m$ that simultaneously enhances the intra-class compactness between normal and augmented-normal frames and inter-class discrepancy between normal and pseudo-abnormal frames.} 
We design $\mathcal{L}_{\text{RAA}}$ using the ArcFace loss \cite{deng2019arcface} enforcing this margin as follows.
\begin{equation}
    \medmuskip=0mu
    \thinmuskip=0mu
    \thickmuskip=0mu
    \hspace*{-0.6em}
    \small{
    \mathcal{L}_{\text{RAA}} = \dfrac{-1}{N}\sum_{i=0}^{N-1}\log\Bigg{(}\dfrac{e^{s(\cos(\omega_{y_i} + m))}}{e^{s(\cos(\omega_{y_i} + m))}+\sum_{j=0,j\neq y_i}^1e^{s\cos(\omega_j)}}\Bigg{)},
    }
    \label{eq:rel_att_loss}
\end{equation}
{where label $y_i$ is set as 1 for normal frame $\widehat{\bm{v}}$ and augmented frame $g(\widehat{\bm{v}})$, and 0 for pseudo-abnormal frame $\tilde{\bm{v}}$. We transform $\mathcal{A}(\bm{x})$ with $\psi_{y_i}= \Vert\bm{W}_{y_i}\Vert\Vert\texttt{vec}(\mathcal{A}(\bm{x}))\Vert\cos(\omega_{y_i})$ (with $\omega_{y_i}\in[0, \pi]$ as the angle between $\bm{W}_{y_i}$ and $\texttt{vec}(\mathcal{A}(\bm{x}))$). Here, $\texttt{vec}(\cdot)$ is a vectorizing operation. $\Vert\bm{W}_{y_i}\Vert$ and $\Vert\texttt{vec}(\mathcal{A}(\bm{x}))\Vert$ are normalized to 1 which leads to $\psi_{y_i}=\cos(\omega_{y_i})$. With ArcFace loss, $\bm{W}_{y_i}$ behaves as a centre for each class (\ie normal and abnormal) \cite{deng2019arcface} which creates a distance margin penalty of $m$. We set scaling factor $s$ = 64 and margin $m$ = 28.6 degrees following \cite{musgrave2020pytorch}. $\mathcal{L}_{\text{RAA}}$ can be implemented as any triplet metric learning loss \cite{musgrave2020pytorch}. However, we choose the ArcFace loss as it has been shown to perform well in recent non-VAD works \cite{wu2021learning, yang2021dolg, afrasiyabi2021mixture}.}

\paragraph{Final learning objectives.} To summarize, \zvad is trained end-to-end with $\mathcal{G}(\cdot)$ learning loss $\mathcal{L}_{\mathcal{G}}$, $\mathcal{D}(\cdot)$ learning loss $\mathcal{L}_{\mathcal{D}}$, and $\mathcal{N}(\cdot)$ learning loss $\mathcal{L}_{\mathcal{N}}$ as follows:
\begin{align}
\begin{gathered}
    \mathcal{L}_{\mathcal{G}} = \mathcal{L}_{\text{BB}} +
    \alpha_{\mathcal{D}}\mathbb{E}_{\widehat{\bm{v}}\sim \rho}\big{[}\nicefrac{1}{2}\big{(}\mathcal{D}(\widehat{\bm{v}}) - 1\big{)}^2\big{]} + \\ \alpha_{\mathcal{N}}\mathbb{E}_{\widehat{\bm{v}}\sim \rho}\big{[}\nicefrac{1}{2}\big{(}\mathcal{N}(\widehat{\bm{v}}) - 1\big{)}^2\big{]},\\
    \hspace{-0.8em}\mathcal{L}_{\mathcal{D}} = \mathbb{E}_{\widehat{\bm{v}}\sim \rho}\big{[}\nicefrac{1}{2}\big{(}\mathcal{D}(\widehat{\bm{v}})\big{)}^2\big{]} + \mathbb{E}_{\widehat{\bm{v}}\sim \rho}\big{[}\nicefrac{1}{2}\big{(}\mathcal{D}(\bm{v}) - 1\big{)}^2\big{]},\\
    \mathcal{L}_{\mathcal{N}} = \alpha_{\text{n}}\mathcal{L}_{\text{N}} + \alpha_{\text{rn}}\mathcal{L}_{\text{RN}} + \alpha_{\text{aa}}\mathcal{L}_{\text{AA}} +
    \alpha_{\text{raa}}\mathcal{L}_{\text{RAA}}~~~~ 
\end{gathered}
\end{align}
We set $\alpha_{\mathcal{D}} = 0.05$ following \cite{liu2018future}. The rest of loss weights $\alpha_{\mathcal{N}} = 0.5, \alpha_{\text{n}} = 1, \alpha_{\text{rn}}=0.01, \alpha_{\text{aa}}=1$, and $\alpha_{\text{raa}}=1$ are set empirically.
\paragraph{Discussion on why} \zvad \textbf{works.} \zvad trains $\mathcal{G}(\cdot)$ in predicting the future-frame $\widehat{\bm{v}}$ of input normal video by considering the difference with respect to pseudo-abnormal frames (\textit{via} our normalcy classifier module). Not considered in prior works, this strategy specifically helps $\mathcal{G}(\cdot)$ to learn contextual relative difference between normal and abnormal frames to alleviate overfitting to source domain normal video features. The overfitting issue is further mitigated when the abnormal examples created from our pseudo-anomaly module contains various kinds of objects as ``foreign entities" in VAD normal frames. This allows $\mathcal{G}(\cdot)$ to learn the relative normalcy difference from extremely diverse kinds of pseudo-anomalies, making it capable of detecting different anomaly types (in inference-time) in multiple target domains without any prior knowledge.

To make the further discussion concise, we show the statistics and acronyms of the VAD and TI datasets in Tab. \ref{tab:dataset_stats}.

\subsection{Introduction to Task-Irrelevant (TI) Datasets}
\label{sec:TI_dataset}
\begin{table*}[t]
\centering
\setlength{\tabcolsep}{2pt}
\caption{\textbf{Dataset statistics.} We highlight the difference in amount of training data between VAD and TI datasets. $^\star$: the train/test disjoint camera (dc) split is provided by \cite{lu2020few}. As stated in \cite{lu2020few}, UCFC dataset does not contain ground truth frame-level labels and hence is not considered for evaluation.}
%\vspace{-.5em}
\resizebox{0.95\textwidth}{!}{%
\begin{tabular}{@{}c@{}cccccccccc@{}}
\toprule
 & \multicolumn{6}{c}{\textbf{Task-Relevant/VAD Datasets}} & & \multicolumn{3}{c}{\textbf{Task-Irrelevant/non-VAD Datasets}} \\
\cmidrule(l){2-7} \cmidrule(l){8-11} 
\multirow{-1}{*}{\textbf{Property} $\backslash$ \textbf{Dataset}} & Shanghai-Tech \cite{luo2017revisit} & Shanghai-Tech$^\star$ \cite{lu2020few} & UCF-Crime \cite{sultani2018real} & Ped1 \cite{mahadevan2010anomaly} & Ped2 \cite{mahadevan2010anomaly} & CUHK-Avenue \cite{lu2013abnormal} & & HMDB51 \cite{kuehne2011hmdb} & UCF101 \cite{soomro2012ucf101} & 20BN-JESTER \cite{materzynska2019jester}\\
\midrule
Acronym & SHT & SHT$_{\text{dc}}$ & UCFC & Ped1 & Ped2 & Ave & & HMDB & UCF101 & Jester\\
\# of training / testing videos & 330 / 107 & 147 / 33 & 950 / - & 34 / 36 & 16 / 12 & 16 / 21 & & 6,766 / - & 13,320 / - & 50,420 / -\\
\# of abnormal instances & 47 & 33 & -- & 40 & 12 & 21 & & N/A & N/A & N/A\\
\bottomrule
\end{tabular}%
}
%\vspace{-0.5em}
\label{tab:dataset_stats}
\vspace*{-\baselineskip}
\end{table*}

\noindent In this section, we discuss the utilities of task-irrelevant or non-VAD videos for unsupervised VAD. Task-relevant or VAD datasets provided by VAD research community are known to be limited in scale as shown in Tab. \ref{tab:dataset_stats} and \cite{wananomaly, acsintoae2022ubnormal, lu2020few}. (\eg Ave \cite{lu2013abnormal}, Ped1, Ped2 \cite{mahadevan2010anomaly} datasets have $< 100$ training videos). Further, it is difficult to collect different kinds of scenarios of normal activities with such limited scale. Hence, we propose to introduce the utility of Task-Irrelevant (TI) datasets to the task of VAD. 

We define a dataset as `\textit{Task-Irrelevant}' which is freely available from different other video downstream or non-VAD tasks (\eg, video classification, action recognition, \etc.). Examples of such datasets are UCF101 \cite{soomro2012ucf101} and HMDB \cite{kuehne2011hmdb} (see Tab.~\ref{tab:dataset_stats}). Such datasets were originally introduced for non-VAD works, specifically curated for large-scale deep learning-based tasks. For example, Jester was originally introduced for video classification of 25 hand gesture classes \cite{materzynska2019jester}. To show the performance using diverse types of datasets in our \zvad task, we choose Jester, UCF101, and HMDB to be our TI datasets. Please see Supplementary Material for dataset examples. Next, we discuss how the task-relevancy of these datasets is measured with respect to the VAD task, followed by two simple strategies to use these datasets in the proposed problem scenario. {Note that \zvad needs nothing from the TI-VAD relevancy measure to operate. The purpose is to only validate TI data's irrelevancy to the VAD task.}
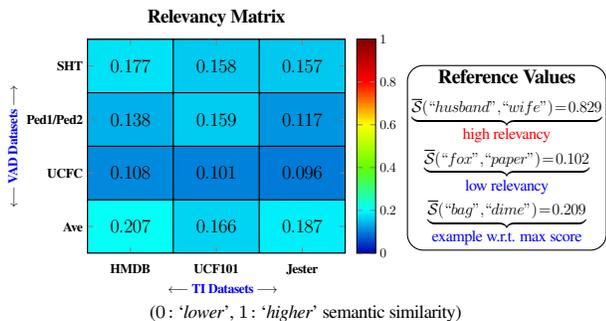
\begin{figure}[t]
    %\begin{minipage}[c]{0.5\columnwidth}
    \raggedright
    \begin{tikzpicture}[scale=0.5, transform
        shape]
        \begin{axis}[
            name=relMat,
            % view={0}{90},   % not needed for `matrix plot*' variant
            xlabel= {~~~~$\longleftarrow$ \textcolor{blue}{\textbf{TI Datasets}} $\longrightarrow$},
            ylabel= $\longleftarrow$ \textcolor{blue}{\textbf{VAD Datasets}} $\longrightarrow$,
            colormap/bluered,
            colorbar,
            colorbar style={
                % title=\%,
                samples = 3,
                yticklabel style={
                    /pgf/number format/.cd,
                    fixed
                },
            },
            title={\Large \bf Relevancy Matrix},
            %
            % added these key-values
            enlargelimits=false,
            axis on top,
            nodes near coords,
            nodes near coords style={
                font=\bfseries,
                anchor=center,
                font=\Large,
                /pgf/number format/fixed,
                /pgf/number format/precision=3
            },
            point meta min=0,
            point meta max=1,
            xticklabels={HMDB, UCF101, Jester},
            xtick={0,...,2},
            yticklabels={Ave, UCFC, Ped1/Ped2, SHT},
            ytick={0,...,3},
            x tick label style={yshift=-0.5em},
            tick label style={font=\bfseries},
            legend style={at={(0.5,-0.25)},anchor=north}
        ]
    
            \addplot [matrix plot*, draw = black,
            point meta=explicit, no markers] file {relevancy-matrix.dat};
    
        \end{axis}
        %\hspace*{-1em}
        \node[below right, align=center, text=black, 
            draw,
            fill=white,
            rounded corners=6pt,]
            % blur shadow={shadow blur steps=5}]
            at (8.5, 5) {%
            \fontsize{15}{15}\selectfont
            \textbf{Reference Values} \\[1em]
            $\begin{gathered} 
            \fontsize{12}{12}\selectfont
            \underbrace{\overline{\mathcal{S}}(``husband", ``wife") = 0.829}_{\textrm{\textcolor{red}{\fontsize{12}{12}\selectfont high relevancy}}}\\
            \fontsize{12}{12}\selectfont
            \underbrace{\overline{\mathcal{S}}(``fox", ``paper") = 0.102}_\textrm{\textcolor{blue}{\fontsize{12}{12}\selectfont
            low relevancy}}\\
            \fontsize{12}{12}\selectfont
            \underbrace{\overline{\mathcal{S}}(``bag", ``dime") = 0.209}_\textrm{\textcolor{blue}{\fontsize{12}{12}\selectfont
            example w.r.t. max score}}
            \end{gathered}$};
    \end{tikzpicture}
    \centering
    \scriptsize{ ($0:$ `\textit{lower}', $1:$ `\textit{higher}' semantic similarity)}
    %\vspace{-1.8em}
    \caption{\textbf{Relevancy measure between VAD and TI labels.} Using the relevancy score matrix between the TI and VAD labels, we find that TI datasets have low semantic similarity with the VAD datasets. The maximum score occurs between HMDB and Ave.}
    \label{fig:mean-heatmap}
  %\end{minipage}
\vspace*{-1.25\baselineskip}
\end{figure}
\paragraph{Measuring relevancy of non-VAD datasets.} Following \cite{peng2018zero,fu2015zero}, we use word2vec \cite{mikolov2013distributed} (pre-trained on Google News dataset \cite{mikolov2013distributed}) to measure the task-relevancy of our introduced TI datasets: Jester, UCF101, and HMDB. We first compute an embedding vector of the input labels (in case the label contains more than one word, we average the embedding). Next, we compute the mean absolute cosine similarity $\overline{\mathcal{S}}\in[0,1]$ of the embedding for all possible pairs of labels between the TI datasets and abnormal classes of VAD datasets. This is denoted as $\overline{\mathcal{S}} = \nicefrac{1}{\ell_P\ell_Q}\sum_{p=1}^{\ell_P}\sum_{q=1}^{\ell_Q}\big{\vert}\text{cos-sim}\big{(}\bm{\pi}_p, \bm{\pi}_q\big{)}\big{\vert},$
where $\ell_P$ and $\ell_Q$ are the total number of labels in the TI and VAD dataset, $\bm{\pi}_p$ and $\bm{\pi}_q$ are the word2vec representation of the $p$-th and $q$-th label of the TI and VAD dataset, respectively. $\text{cos-sim}(\cdot)$ denotes the cosine similarity operation on input vectors. A value of $\overline{\mathcal{S}}$ closer to 0 indicates a higher degree of irrelevancy (or lower degree of relevance). In Fig. \ref{fig:mean-heatmap}, we show the mean cosine similarity $\overline{\mathcal{S}}$ for all the TI (\ie, HMDB, UCF101, Jester) and VAD (\ie, SHT, Ped1/Ped2, UCFC, Ave) datasets used in this paper. Following reference values:
$\overline{\mathcal{S}}(``object", ``scene")=0.829$, $\overline{\mathcal{S}}(``bag", ``dime")=0.209$, and $\overline{\mathcal{S}}(``fox", ``paper")=0.109$, we find
that the maximum semantic similarity $\overline{\mathcal{S}}=0.207$ occurs between Ave and HMDB indicating that all the TI datasets are quite irrelevant to the task of VAD problem.

\paragraph{Methods to use TI datasets.} We provide two methods to use TI datasets. \textit{Firstly}, unsupervised VAD methods learn features from normal events during training. These events are particularly marked by continuous activities without any sudden disruption from alien objects. Such kinds of videos are readily available in other video downstream tasks like action recognition, where a sample video only contains frames from a continuous activity. In cases where there is no VAD training data available in the source domain (worst-case scenario), we show later in Sec. \ref{sec:experiments} that training \zvad solely with TI datasets reaches the SOTA results across 3 different target domain datasets. We hypothesize that TI datasets represent the recording of normal activities as in VAD training data with normal videos. Hence, learning from such TI data helps in modeling features similar to normal videos. \textit{Secondly}, we recommend using TI frames to create anomalies containing diverse types of objects (see Fig. \ref{fig:pseudo_anomaly}). Using our proposed method to create pseudo-anomaly frames using TI data (details in Sec. \ref{sec:pseudo_anomaly}), our generator learns features from normal frames \textit{relative} to abnormal frames. Such pseudo-anomalies contain diverse foreign entities extracted from TI video frames allowing our generator to learn relative normalcy difference in a broad manner.

%--------- EXPERIMENTATION ----------% 
\section{Experiments and Results}\label{sec:experiments}
\vspace{0.8em}
 
\paragraph{Implementation details.} We implement our framework in PyTorch \cite{paszke2019pytorch}. The generator $\mathcal{G}(\cdot)$ is an U-Net \cite{ronneberger2015u} adapted from \cite{liu2018future} with a memory module at its bottleneck similar to \cite{dong2020dual}. The discriminator $\mathcal{D}(\cdot)$ and normalcy classifier $\mathcal{N}(\cdot)$ are Patch-GAN discriminators \cite{isola2017image}. We provide more details of our implementation in Supplementary Material. \\
\vspace{0.0005em}

\noindent\textbf{Evaluation details.} We evaluate \zvad under three training scenarios with respect to types of available source data: (1) \textit{Both VAD and TI data are available}: $\mathcal{G}(\cdot)$ takes VAD videos and $\modO$ takes TI frames as input, (2) \textit{Only one of the VAD or TI data are available}: Both $\mathcal{G}(\cdot)$ and $\modO$ take VAD or TI videos as input. We did not observe any performance gain empirically when $\mathcal{G}(\cdot)$ takes both VAD and TI videos as input, so we drop this case as it adds a computational burden. \tblue{We compare \zvad with \cite{lu2020few, lv2021learning} using the area under ROC curve (AUC), model storage, total parameters, GPU energy consumption, inference time FPS, and GMACs.} \\
\vspace{0.0005em}

\noindent\textbf{Baselines.} Since the problem of `cross-domain VAD without target domain training data adaptation' is identified by us, we cannot find other methods which are designed for such a setup. The latest and closest baselines we found are \rgan and \mpn, which are designed for the xVAD task without needing strong priors from VAD frame object extraction. Since both methods report their performance under the proposed problem setup, we use them as our baselines. Even though we outperform strong prior based xVAD methods \cite{roy2021predicting, georgescu2020background, ionescu2019object} without any such computationally expensive operation under our problem setup, we do not consider them as part of our baselines for fair comparisons with respect to \cite{lu2020few, lv2021learning}. In Tab. \ref{tab:same_dataset_exp}, and \ref{tab:cross_domain_testing}, \textit{paper} denotes results as reported and \textit{code} denotes results computed using official code, if available.

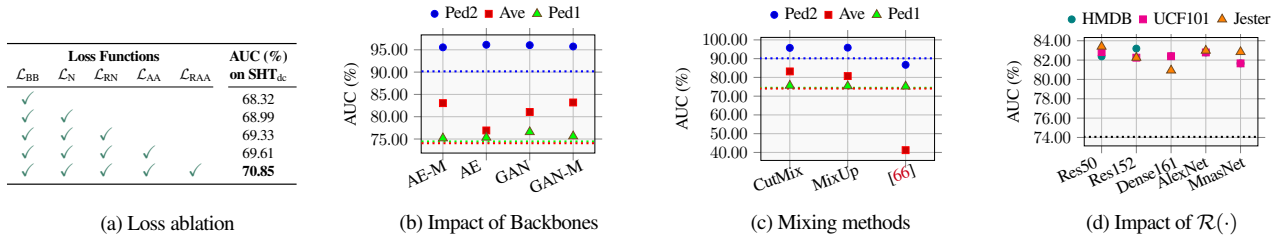
\begin{figure*}[t!]
    \vspace*{-3\baselineskip}
    \centering
    \begin{subfigure}[t]{0.245\textwidth}
        \raggedright
        \vspace*{-6em}
        \resizebox{0.9\columnwidth}{!}{%
\begin{tabular}{cccccc}
\toprule
\multicolumn{5}{c}{\textbf{Loss Functions}} &\textbf{AUC (\%)}\\
$\mathcal{L}_{\text{BB}}$ & $\mathcal{L}_{\text{N}}$ & $\mathcal{L}_{\text{RN}}$ & $\mathcal{L}_{\text{AA}}$ & $\mathcal{L}_{\text{RAA}}$ &\textbf{on SHT}$_{\text{dc}}$ \\ \cmidrule(r){1-5} \cmidrule(l){6-6} 
$\ccheck$ &  &  &  &  & 68.32 \\
$\ccheck$ & $\ccheck$ &  &  &  &  68.99\\
$\ccheck$ & $\ccheck$ & $\ccheck$ &  &  &  69.33\\
$\ccheck$ & $\ccheck$ & $\ccheck$ & $\ccheck$ &  &  69.61 \\
$\ccheck$ & $\ccheck$ & $\ccheck$ & $\ccheck$ & $\ccheck$ &  \textbf{70.85}\\ \bottomrule\\
\end{tabular}%
}
        \vspace{-0.2\baselineskip}
        \caption{Loss ablation}
        \label{tab:ablation}
    \end{subfigure}%
    ~
    \begin{subfigure}[t]{0.245\textwidth}
        \raggedright
        \pgfplotsset{every x tick label/.append style={font=\normalsize, rotate=25, xshift=-1.5ex, yshift=2ex}}

\pgfplotsset{y tick label style={/pgf/number format/fixed zerofill, /pgf/number format/precision=2, font=\normalsize}}

\pgfplotstableread[col sep=&, header=true]{
  backbone      & ped   & ped_  & ave
  AE-M          & 95.55 & 75.17 & 83.04  
  AE            & 96.13 & 75.34 & 76.92  
  GAN           & 96.04 & 76.57 & 81.06  
  GAN-M         & 95.74 & 75.62 & 83.19  
}\bb % backbone impact (ped2, avenue, ped1)

\begin{tikzpicture}[scale=0.65, transform shape, every node/.append style={font=\normalsize}]
    \begin{axis}
      [
        xtick=data,
        xticklabels from table={\bb}{backbone},  
        width= 1.2\textwidth,
        height = 40mm,
        ytick={70, 75, ...,100},
        nodes near coords align={vertical},
        xmin=0.5, xmax=4.5,
        ylabel={AUC (\%)},
        legend style={at={(0.5,1.30)}, anchor=north, column sep=0.5ex, draw=none, only marks, font=\norma},
        legend columns=-1,
        axis background/.style={fill=gray!5},
        grid=both,
        ytick align=outside,
        xtick align=outside
      ]
      \addplot table [y=ped, x expr=\coordindex+1, only marks] {\bb};
      \addlegendentry{Ped2}
      \addplot table [y=ave, x expr=\coordindex+1, only marks] {\bb};
      \addlegendentry{Ave}
      \addplot+[mark=triangle*, mark size=3pt, mark options={fill=green}, only marks] table [y=ped_, x expr=\coordindex+1] {\bb};
      \addlegendentry{Ped1}
      \addplot[mark=none, very thick, dotted, green, domain=0:6] {74.45};
      \addplot[mark=none, very thick, dotted, blue, domain=0:6] {90.17};
      \addplot[mark=none, very thick, dotted, red, domain=0:6] {74.06};
    \end{axis}
  \end{tikzpicture}
        \vspace{-0.2\baselineskip}
        \caption{Impact of Backbones}
        \label{fig:backbone_nw_strategy}
    \end{subfigure}%
    ~ 
    \begin{subfigure}[t]{0.245\textwidth}
        \raggedright
        \pgfplotsset{every x tick label/.append style={font=\normalsize, rotate=20, xshift=-0.9ex, yshift=2ex}}

\pgfplotsset{y tick label style={/pgf/number format/fixed zerofill, /pgf/number format/precision=2, font=\normalsize}}

\pgfplotstableread[col sep=&, header=true]{
  method                               & ped   & ave   & ped_
  CutMix                       & 95.74 & 83.19 & 75.62
  MixUp                      & 95.86 & 80.70 & 75.34
  %None                         & 95.55 & 83.03 & 76.64
  {~~~~\cite{astrid2021learning}}           & 86.70 & 41.20 & 75.18
}\mix % pseudo-anomaly strategy (ped2, avenue, ped1)

\begin{tikzpicture}[scale=0.65, transform shape, every node/.append style={font=\normalsize}]
    \begin{axis}
      [
        xtick=data,
        xticklabels from table={\mix}{method},  
        width = 1.2\textwidth,
        height = 41mm,
        ytick={40,50,...,100},
        nodes near coords align={vertical},
        xmin=0.5, xmax=3.5,
        ylabel={AUC (\%)},
        legend style={at={(0.5,1.3)}, anchor=north, column sep=0.5ex, draw=none, only marks},
        legend columns=-1,
        axis background/.style={fill=gray!5},
        grid=both,
        ytick align=outside,
        xtick align=outside
      ]
      \addplot table [y=ped, x expr=\coordindex+1, only marks] {\mix};
      \addlegendentry{Ped2}
      \addplot table [y=ave, x expr=\coordindex+1, only marks] {\mix};
      \addlegendentry{Ave}
      \addplot+[mark=triangle*, mark size=3pt, mark options={fill=green}, only marks] table [y=ped_, x expr=\coordindex+1] {\mix};
      \addlegendentry{Ped1}
      \addplot[mark=none, very thick, dotted, green, domain=0:6] {74.45};
      \addplot[mark=none, very thick, dotted, blue, domain=0:6] {90.17};
      \addplot[mark=none, very thick, dotted, red, domain=0:6] {74.06};
    \end{axis}
  \end{tikzpicture}
        \vspace{-0.2\baselineskip}
        \caption{Mixing methods}
        \label{fig:pseudo_ano_strategy}
    \end{subfigure}%
    ~ 
    \begin{subfigure}[t]{0.245\textwidth}
        \raggedright
        \pgfplotsset{every x tick label/.append style={font=\normalsize, rotate=25, xshift=-2.5ex, yshift=2ex}}

\pgfplotsset{y tick label style={/pgf/number format/fixed zerofill, /pgf/number format/precision=2, font=\normalsize}}

\pgfplotstableread[col sep=&, header=true]{
  network      & hmdb   & ucf    & jester
  Res50     & 82.38  & 82.83  & 83.40
  Res152    & 83.19  & 82.25  & 82.25
  Dense161  & 82.40  & 82.40  & 80.93
  AlexNet   & 82.79  & 82.79  & 82.98
  MnasNet   & 81.66  & 81.66  & 82.84
}\avenue % avenue

  \begin{tikzpicture}[scale=0.65, transform shape, every node/.append style={font=\normalsize}]
    \begin{axis}
      [
        xtick=data,
        xticklabels from table={\avenue}{network},      
        width = 1.2\textwidth,
        height = 38mm,
        ytick={74,76,...,84},
        nodes near coords align={vertical},
        xmin=0.5, xmax=5.5,
        ylabel={AUC (\%)},
        legend style={at={(0.5,1.3)}, anchor=north, column sep=0.5ex, draw=none, only marks},
        legend columns=-1,
        axis background/.style={fill=gray!5},
        grid=both,
        ytick align=outside,
        xtick align=outside
      ]
      \addplot+[color=teal, mark options={fill=teal}] table [y=hmdb, x expr=\coordindex+1, only marks] {\avenue};
      \addlegendentry{HMDB}
      \addplot+[color=magenta, mark options={fill=magenta}] table [y=ucf, x expr=\coordindex+1, only marks] {\avenue};
      \addlegendentry{UCF101}
      \addplot+[mark=triangle*, mark size=3pt, mark options={fill=orange}, only marks] table [y=jester, x expr=\coordindex+1] {\avenue};
      \addlegendentry{Jester}
      \addplot[mark=none, very thick, dotted, black, domain=0:6] {74.06};
    \end{axis}
  \end{tikzpicture}
        \vspace{-0.2\baselineskip}
        \caption{Impact of $\mathcal{R}(\cdot)$}
        \label{fig:rand_nw_strategy}
    \end{subfigure}%
    %\vspace{-.5em}
    \caption{\textbf{Component Analysis of} \zvad. Fig. \ref{tab:ablation} shows the loss ablation using SHT$_{\text{dc}}$; Fig. \ref{fig:backbone_nw_strategy} compares the cross-domain performance of \zvad with different future-frame prediction backbones on three datasets (source: SHT); Fig. \ref{fig:pseudo_ano_strategy} compares the impact of different mixing strategies in module $\modO$ with SOTA method \cite{astrid2021learning} that also presents a pseudo-anomaly method on three datasets (source: SHT); Fig. \ref{fig:rand_nw_strategy} compares the impact of network $\mathcal{R}(\cdot)$ in module $\modO$ on three TI datasets (source: SHT, target: Ave). \textit{Dotted} lines in Fig. \ref{fig:backbone_nw_strategy}, \ref{fig:pseudo_ano_strategy} (three datasets), and \ref{fig:rand_nw_strategy} (one dataset) show SOTA (\mpn with Ped1: 74.45\%, Ped2: 90.17\%, Ave: 74.06\%) in respective cross-domain VAD when source is SHT.}
    \label{fig:component_analysis}
    \vspace*{-0.5\baselineskip}
\end{figure*}
\begin{table}[t]
\caption{\textbf{Comparison in Efficiency and Same-dataset testing.} We beat our baselines in most of the same-dataset testing, and outperform them in the listed efficiency metrics. $^\star$: GPU energy consumption is measured by testing on Ped2. $^\dagger$: \rgan does not provide its official testing code for inference-time metric evaluation.
}
%\vspace{-.5em}
\resizebox{\columnwidth}{!}{%
\begin{tabular}{cccccccccc}
\toprule
 & \multicolumn{5}{c}{\textbf{Efficiency Metrics}} & & \multicolumn{3}{c}{\textbf{Same Dataset Testing}} \\
\cmidrule(l){2-6} \cmidrule(l){7-10} 
\multirow{-2}{*}{\textbf{Method}} & \begin{tabular}[c]{@{}c@{}}\textbf{Parameters} ($\down$)\\ (millions)\end{tabular} &\textbf{GMACs} ($\down$) & \begin{tabular}[c]{@{}c@{}}\textbf{Energy} ($\down$)\\ (Joules)$^\star$\end{tabular} & \begin{tabular}[c]{@{}c@{}}\textbf{Storage} ($\down$)\\ (MegaByte)\end{tabular} & \textbf{FPS} ($\up$) & & \textbf{SHT}$_{\text{dc}}$ & \textbf{Ped2} & \textbf{SHT} \\
\midrule
\rgan & 19.0 & 1384.52 & --$^\dagger$ & 79.85 & 2.1 & & 70.11 & 96.90 & \textbf{77.90} \\
\mpn  & 12.7 & 55.09 & 10.65 & 53.14 & 166.8 & & 67.47 & 96.20 & 73.80 \\
\rowcolor{red!10}
\zvad & \textbf{8.73} & \textbf{43.10} & \textbf{6.81} & \textbf{34.92} & \textbf{208.5} & & \textbf{70.85} & \textbf{96.95} & 71.60 \\
\bottomrule
\end{tabular}%
}
\label{tab:same_dataset_exp}
\vspace{-\baselineskip}
\end{table}

\begin{table*}[t]
\begin{minipage}[!t]{0.495\textwidth}
\centering
\raggedright
\captionsetup{width=0.985\columnwidth}
\caption{\textbf{Cross-dataset testing.} Comparison with xVAD works that need no background-subtraction. The best and second best AUC are marked in \textbf{bold} and \underline{underline}, respectively. $^\ddagger$: For \mpn, the publicized code~\cite{MPNcode} gives lower AUC than what was reported in their paper.
%`N/A' denotes `Not Applicable.'
}
%\vspace*{-0.5em}
\resizebox{0.975\columnwidth}{!}{%
\begin{tabular}{cccccc}
\toprule
& &  & \multicolumn{3}{c}{\textbf{VAD Testing Data}} \\
 \cmidrule(l){4-6} 
\multirow{-2}{*}{\dual{\textbf{VAD Training Data}/(\textbf{Input to $\mathcal{G}(\cdot)$})}} &\multirow{-2}{*}{\dual{\textbf{Auxiliary Data}/(\textbf{Input to $\modO$})}} &\multirow{-2}{*}{\textbf{Method}} &\textbf{Ped1} &\textbf{Ped2} &\textbf{Ave} \\
\midrule
SHT &N/A &\rgan (paper) &73.10 &81.95 &71.43  \\
SHT &N/A &\mpn (paper) &74.45 &90.17 &74.06  \\
SHT &N/A &\mpn (code)$^\ddagger$ &66.05 &84.73 &74.06  \\
\rowcolor{red!10}
SHT &SHT &\zvado  &\textbf{76.14} &\underline{95.78} &82.28 \\
\rowcolor{red!10}
SHT &HMDB &\zvado &75.62 &95.74 &\textbf{83.19} \\
\rowcolor{red!10}
SHT &UCF101 &\zvado &75.41 &\textbf{95.80} &82.25 \\
\rowcolor{red!10}
SHT &Jester &\zvado &\underline{75.93} &95.62 &\underline{82.49} \\
\midrule
UCFC &N/A &\rgan (paper) &66.87 &62.53 &64.32 \\
UCFC &N/A &\mpn (paper) &75.52 &86.04 &\textbf{82.26} \\
\rowcolor{red!10}
UCFC &UCFC &\zvado &\textbf{78.61} &\textbf{91.65} & 81.11 \\
\rowcolor{red!10}
UCFC &HMDB &\zvado & 78.02 & 87.66 & 81.50 \\
\rowcolor{red!10}
UCFC &UCF101 &\zvado & 76.27 & 86.80 & 81.45 \\
\rowcolor{red!10}
UCFC &Jester &\zvado &\underline{78.39} &\underline{88.71} &\underline{81.55} \\
\bottomrule
\end{tabular}%
}
\label{tab:cross_domain_testing}
\vspace*{0.5em}
\raggedright
\captionsetup{width=0.98\columnwidth}
\caption{\textbf{Cross-dataset testing.} Cross-domain performance when \zvad model only trained with TI datasets. 
}
\label{tab:ti_cross_domain_testing}
\resizebox{0.975\columnwidth}{!}{%
\begin{tabular}{ccccc}
\toprule
& & \multicolumn{3}{c}{\textbf{VAD Testing Data}} \\
 \cmidrule(l){3-5} 
\multirow{-2}{*}{\dual{\textbf{TI Training Data}/(\textbf{Input to $\mathcal{G}(\cdot)$})}} &\multirow{-2}{*}{\dual{\textbf{Auxiliary Data}/(\textbf{Input to $\modO$})}} &\textbf{Ped1} &\textbf{Ped2} &\textbf{Ave} \\
\midrule
HMDB &HMDB &\underline{76.66} & \textbf{91.53} &\textbf{81.92} \\
UCF101 &UCF101 & {75.67} & {85.84} & \underline{81.78} \\
Jester &Jester & \textbf{78.12} &\underline{91.23} & {78.06} \\
\bottomrule
\end{tabular}%
}
\end{minipage}%
\hfill
\begin{minipage}[!t]{0.495\textwidth}
    \vspace{-7em}
    \centering
    \raggedleft
    \pgfplotsset{y tick label style={/pgf/number format/fixed zerofill, /pgf/number format/precision=2, font=\Large}}
\begin{tikzpicture}[scale=0.3, transform shape, every node/.append style={font=\Large}]
    \begin{axis}[
        ybar,
        bar width = 0.5,
        enlargelimits=0.15,
        ylabel={AUC (\%)},
        xtick={1,2,3,4},
        ytick={70,72,...,78},
        title style={at={(0.35,0.875)},anchor=north, font=\Large},
        title = SHT $\rightarrow$ Ped1,
        title style={fill=yellow!30},
        xticklabels={\cite{roy2021predicting},\cite{ionescu2019object},\cite{georgescu2020background},\zvad},
        width={\textwidth}, 
        nodes near coords,
        nodes near coords align={vertical},
        axis background/.style={fill=gray!5},
        ytick align=outside,
        xtick align=outside,
        ymax=78, ymin=70,
        ytick pos=left,
        xtick pos=left,
        xticklabel style={font=\Large, rotate=0, xshift=0ex, yshift=0ex}
        ]
        \addplot[fill=other-color, bar shift=0mm] coordinates {(1,71.60) (2,74.40)};
        \addplot[fill=other-color, bar shift=0mm, nodes near coords={}] coordinates {(3,69)} node[pos=1,above,fill=gray!5] {N/P};
        \addplot[fill=zvad-color, mark=no, bar shift=0mm] coordinates {
         ((4,76.14)};
    \end{axis}
\end{tikzpicture}
\begin{tikzpicture}[scale=0.3, transform shape, every node/.append style={font=\Large}]
    \begin{axis}[
        ybar,
        bar width = 0.5,
        enlargelimits=0.15,
        ylabel={AUC (\%)},
        xtick={1,2,3,4},
        ytick={74,76,...,84},
        title style={at={(0.35,0.875)},anchor=north},
        title = SHT $\rightarrow$ Ave,
        title style={fill=yellow!30},
        xticklabels={\cite{roy2021predicting},\cite{ionescu2019object},\cite{georgescu2020background},\zvad},
        width={\textwidth}, 
        nodes near coords, 
        nodes near coords align={vertical},
        axis background/.style={fill=gray!5},
        ytick align=outside,
        xtick align=outside,
        ymax=84, ymin=74,
        xtick pos=left,
        ytick pos=left,
        x tick label/.append style={font=\Large, rotate=0, xshift=0ex, yshift=0ex}
        ]
        ]
        \addplot[fill=other-color, bar shift=0mm] coordinates {(1,74.70) (2,78.80) (3,81.00)};
        \addplot[fill=zvad-color, bar shift=0mm] coordinates {(4,82.28)};
    \end{axis}
\end{tikzpicture}
\begin{tikzpicture}[scale=0.3, transform shape, every node/.append style={font=\Large}]
    \begin{axis}[
        ybar,
        bar width = 0.5,
        enlargelimits=0.15,
        ylabel={AUC (\%)},
        xtick={1,2,3,4},
        ytick={90,92,...,100},
        title style={at={(0.35,0.875)},anchor=north},
        title = SHT $\rightarrow$ Ped2,
        title style={fill=yellow!30},
        xticklabels={\cite{roy2021predicting},\cite{ionescu2019object},\cite{georgescu2020background},\zvad},
        width={\textwidth}, 
        nodes near coords, 
        nodes near coords align={vertical},
        axis background/.style={fill=gray!5},
        ytick align=outside,
        xtick align=outside,
        ymax=100, ymin=90,
        xtick pos=left,
        ytick pos=left,
        x tick label/.append style={font=\Large, rotate=0, xshift=0ex, yshift=0ex}
        ]
        \addplot[fill=other-color, bar shift=0mm] coordinates {(2,90.30) (3,95.70) (4,95.78)};
        \addplot[fill=zvad-color, bar shift=0mm] coordinates {(1,95.90)};
    \end{axis}
\end{tikzpicture}
    %\vspace{-0.15em}
    \raggedleft
    \captionsetup{width=0.925\columnwidth}
    
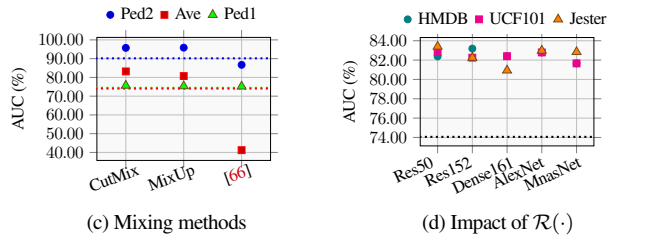
\captionof{figure}{\textbf{Cross-dataset testing.} Comparison with VAD works that need background-subtraction. The best AUC is marked in \textcolor{zvad-color}{red}. `N/P' in leftmost plot means `Not Provided.'}
    \label{fig:background_subtraction}
    %\vspace{0.75\baselineskip}
    \begin{minipage}[!t]{0.49\columnwidth}
        \pgfplotsset{y tick label style={/pgf/number format/fixed zerofill, /pgf/number format/precision=1, /pgf/number format/fixed}}

\begin{tikzpicture}[scale=0.45, transform shape]
    \begin{axis}[
    width= 2.25\textwidth,
    height=60mm,
    xlabel={Frame Index},
    ylabel={Anomaly Score}, 
    title={Ped1 (video \#03)},
    title style={at={(0.15,0.9)},anchor=north, fill=yellow!30, font=\LARGE},
    ymin=-0.1, ymax=1.1,
    xmin = -5, xmax=200,
    grid=major,
    xtick={0, 25, ..., 200},
    ytick={0, 0.2, ..., 1.0},
    axis background/.style={fill=gray!5},
    legend style={fill opacity=0.95, legend pos=south east, font=\LARGE},
    legend style={nodes={scale=0.75, transform shape}},
    legend cell align={left}]
        \addplot[blue, very thick] table [x expr=\coordindex, y index=0]
          {plots/gt_time_series/ped1_gt_3.txt};
        \addplot[red, very thick, dotted] table [x expr=\coordindex, y index=0]
          {plots/mpn_time_series/ped1_03.txt};
        \addplot[red, very thick] table [x expr=\coordindex, y index=0]
          {plots/ours_time_series/ped1_pred_3.txt};
        \legend{Ground-Truth, \mpn, \zvado}
     \end{axis}
\end{tikzpicture}
    \end{minipage}%
    \begin{minipage}[!t]{0.49\columnwidth}
        \pgfplotsset{y tick label style={/pgf/number format/fixed zerofill, /pgf/number format/precision=1, /pgf/number format/fixed}}

\begin{tikzpicture}[scale=0.45, transform shape]
    \begin{axis}[
    width= 2.25\textwidth,
    height=60mm,
    bar width=40,
    xlabel={Frame Index},
    ylabel={Anomaly Score}, 
    title={Ped2 (video \#04)},
    title style={at={(0.15,0.9)},anchor=north, fill=yellow!30, font=\LARGE},
    ymin=-0.1, ymax=1.1,
    xmin = -5, xmax=180,
    grid=major,
    xtick={0, 25, ..., 175},
    ytick={0, 0.2, ..., 1.0},
    axis background/.style={fill=gray!5},
    legend style={fill opacity=0.95, legend pos=south east, font=\LARGE},
    legend style={nodes={scale=0.75, transform shape}},
    legend cell align={left}]
        \addplot[blue, very thick] table [x expr=\coordindex, y index=0]
          {plots/gt_time_series/ped2_gt_4.txt};
        \addplot[red, very thick, dotted] table [x expr=\coordindex, y index=0]
          {plots/mpn_time_series/ped2_04.txt};
        \addplot[red, very thick] table [x expr=\coordindex, y index=0]
          {plots/ours_time_series/ped2_pred_4.txt};
        \legend{Ground-Truth, \mpn, \zvado}
     \end{axis}
\end{tikzpicture}
    \end{minipage}
    \captionsetup{width=0.925\columnwidth}
    \raggedleft
    %\vspace{-0.75\baselineskip}
    
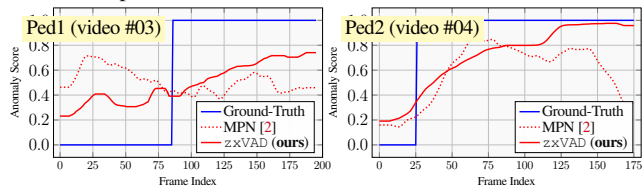
\captionof{figure}{\textbf{Anomaly detection curve.} We compare our cross-domain (source: SHT) anomaly detection on Ped1/Ped2 against \mpn. Larger value in curves indicates possible anomalies.}
    \label{fig:time_series}
\end{minipage}
%\vspace{-0.75\baselineskip}
\end{table*}

\paragraph{Ablation study.} We show the ablation study of our proposed loss functions in \zvad in Fig. \ref{tab:ablation} on the SHT$_{\text{dc}}$ dataset. Fig. \ref{tab:ablation} shows that each of our proposed loss functions contributes to the AUC, and jointly training with them all achieves the best AUC. In Fig. \ref{fig:backbone_nw_strategy}, we analyze different combinations of an autoencoder and generative adversarial network with (AE-M,GAN-M) and without our memory module (AE, GAN) as our \zvad backbone. In Fig.  \ref{fig:pseudo_ano_strategy}, we analyze the impact of different mixing strategies (MixUp \cite{zhang2017mixup}, CutMix \cite{yun2019cutmix} within our module $\modO$ and compare with recent SOTA strategy called Patch \cite{astrid2021learning} that proposes a pseudo-anomaly method. In Fig. \ref{fig:rand_nw_strategy}, we analyze impact of changing $\mathcal{R}(\cdot)$ with (ResNet50, ResNet152 \cite{he2016deep}, DenseNet161 \cite{huang2017densely}, AlexNet \cite{krizhevsky2012imagenet}, MnasNet \cite{tan2019mnasnet}) in \zvad. Fig. \ref{fig:component_analysis} shows that regardless of the backbone choice, the pseudo-anomaly strategy, and the architecture of $\mathcal{R}(\cdot)$, \zvad still outperforms the SOTA baselines in most settings, which supports that \zvad is flexible w.r.t. these factors. 

\noindent\textbf{Same-dataset experiments.}
We compare \zvad with \cite{lu2020few, lv2021learning} on the SHT$_{\text{dc}}$, SHT and Ped2 datasets. Tab. \ref{tab:same_dataset_exp} shows that \zvad outperforms both baselines in AUC in such experiment. For example, \zvad shows better generalization ability across different camera angles than the baselines in the SHT$_{\text{dc}}$ dataset with the least efficiency metrics like model parameters and GMACs. We also find that using extra TI data (HMDB and UCF101) can improve the AUC further compared to baselines (results in Supplementary Material). 

\begin{figure}[!ht]
    \vspace*{-8.75em}
    \centering
    \includegraphics[width=0.87\columnwidth]{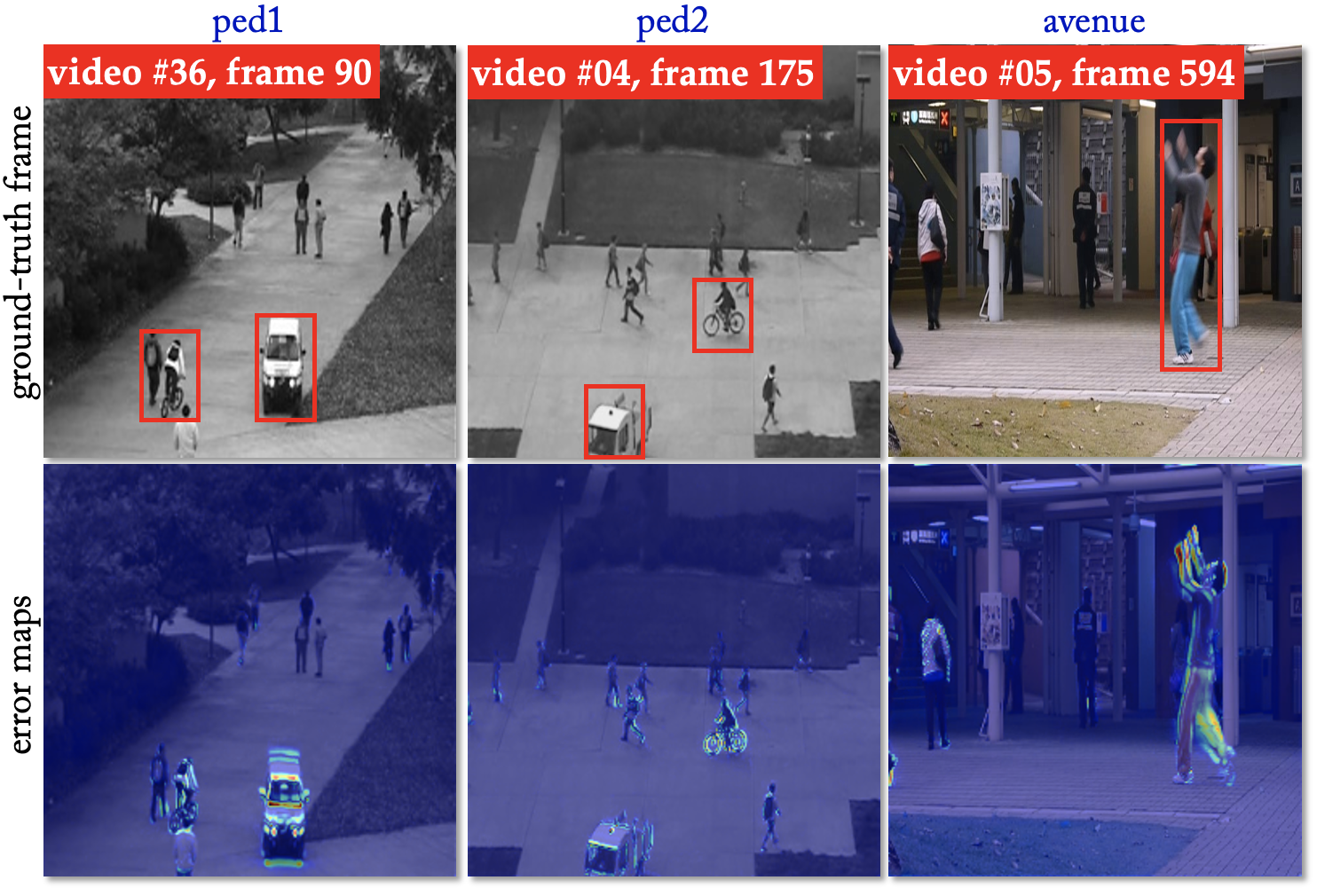}
    \captionof{figure}{\textbf{Difference maps.} We show examples of cross-domain frame prediction comparison on three datasets (source: SHT). The lighter colors in difference map mean larger prediction error indicating anomalies. Red boxes indicate ground truth anomalies.}
    \label{fig:error_maps}
    \vspace*{-1.75em}
\end{figure}

\noindent\textbf{Cross-dataset experiments.} We compare \zvad with \cite{lu2020few, lv2021learning} under the cross-dataset setting. In the top two sections of Tab. \ref{tab:cross_domain_testing}, we train \zvad with either the SHT or UCFC dataset with optional TI data and test it on the Ped1, Ped2, and Ave datasets. Tab. \ref{tab:cross_domain_testing} shows that \zvad outperforms both baselines in AUC under most settings, regardless of whether the extra TI data are used, which supports that \zvad has better generalization ability across different datasets (with different types of anomalies under different scenes) than the baselines. 
For example, when our model is trained on the SHT dataset \cite{luo2017revisit}, it outperforms existing xVAD methods in the proposed problem setup in detecting anomalies like ``chasing" and ``brawling" in SHT's test set as well as anomalies like ``bicycles" and `` cars" in Ped1/Ped2's test set \textit{without} performing any kind of adaptation on Ped1/Ped2's training set. This shows that our method is not specific to anomalies in the source domain, but generalizes well to target domain scenes during inference without adaptation. Tab. \ref{tab:ti_cross_domain_testing} shows that even without using any source domain VAD training data at all, \zvad still outperforms \cite{lu2020few, lv2021learning} in most settings by training with only TI data, which supports our proposed mechanism of using the TI data under the proposed problem setup. These encouraging results suggest that making use of TI data is a promising research direction for the \zvad problem. Interestingly, when either $\mathcal{G}$ \textit{or} $\bm{\mathcal{O}}$ \textit{or} both use TI data, it's not surprising to see slightly lower AUC than if both $\mathcal{G}$ \textit{and} $\bm{\mathcal{O}}$ use VAD relevant data, \ie \textit{more} relevant source data lead to \textit{less} source-target domain gap, resulting in \textit{better} AUC. This is confirmed by average AUC (Tab. \ref{tab:cross_domain_testing} and \ref{tab:ti_cross_domain_testing}) when source is \textit{only} VAD: 84.26\%, \textit{VAD w/ TI}: 83.46\%, and \textit{only} TI: 82.30\%. {We also analyzed the impact of the amount videos needed when solely training with TI data with HMDB and UCF101 in \zvad setup and found that even as little as $\sim$1.25\% of UCF101 or $\sim$8\% of HMDB is enough to outperform the SOTA (details in Supplementary Material).} Following \cite{lu2020few}, we do not perform a cross-domain evaluation with Ped1/Ped2 as a source as the training dataset is too small to make reasonable conclusions. {In Fig. \ref{fig:background_subtraction}, we show that \zvad outperforms existing strong prior based unsupervised xVAD methods \cite{roy2021predicting,georgescu2020background,ionescu2019object} that report cross-domain VAD testing performance when source domain data is SHT. This implies that \zvad provides a computationally efficient and reduced supervision approach with no need for object extraction from videos (using YOLOv3 \cite{redmon2018yolov3} in \cite{georgescu2020background, ionescu2019object} and CenterNet \cite{zhou2019objects} in \cite{roy2021predicting}) both in source and target domain, under the proposed problem setup. Compared to  \cite{georgescu2020background} (in Fig. \ref{fig:background_subtraction}) and \cite{astrid2021learning} (in Fig. \ref{fig:pseudo_ano_strategy}), our untrained CNN based abnormal example generation strategy results in superior VAD for the proposed problem setup.} Our ``relative normalcy'' learning approach optimizes the VAD model to learn features that differentiate normal events from (pseudo)-abnormal events, rather than focusing on learning \textit{only} patterns of normal events as in prior xVAD works. Results in Tab. \ref{tab:ti_cross_domain_testing} (when \zvad uses only TI data) validate this claim as \zvad still outperforms SOTA on target VAD by learning such differentiating features from TI videos. \cite{doshi2022modular} is a few-shot VAD method that puts together three off-the-shelf pre-trained models (YOLOv4 \cite{bochkovskiy2020yolov4}, AlphaPose \cite{fang2017rmpe}, Flownet2 \cite{ilg2017flownet}) to perform xVAD. Even with such costlier storage, high training overhead, and strong priors from different distributions, \zvad easily beats \cite{doshi2022modular} by 11.76\% (Ave), 13.85\% (Ped2) with source as SHT, and 10.12\% (Ave), 29.12\% (Ped2) with source as UCFC with extremely less parameters and no initial priors. Finally, we provide qualitative evaluation under the cross-domain setting with anomaly curves of two testing videos of Ped1 and Ped2 when trained with SHT in Fig.~\ref{fig:time_series}, where \zvad provides better cross-domain detection ability than \mpn. We also visualize difference maps in Fig. \ref{fig:error_maps} (absolute error between ground truth and the predicted frame) that indicate the presence of anomalies by \zvad in three datasets under cross-domain setting after training with SHT. We show more such qualitative results of \zvad in Supplementary Material. In addition to the above, \zvad achieves such results with much better inference-time efficiency than the baselines. Tab. \ref{tab:same_dataset_exp} shows that \zvad outperforms \cite{lu2020few, lv2021learning} in model size, total parameters, GPU energy consumption (computed by pyJoules \cite{pyJoules} following \cite{killamsetty2021retrieve, killamsetty2021grad}), and GMACs by 34.3\%, 31.3\%, 36.1\%, and 21.76\%, respectively.

%------------ CONCLUSION ------------% 
\section{Conclusion}
\noindent We identify a new unsupervised xVAD problem of detecting anomalies in the target domain where no target domain training data are available. To tackle this problem, we propose a novel framework named `Zero-shot Cross-domain Video Anomaly Detection' (\zvad). \zvad aims to learn features of normal activities in input videos by learning how such features are \textit{relatively} different from features of pseudo-abnormal frames. 
Finally, \zvad outperforms the SOTA baselines in most settings under common benchmarks not only in AUC but also in four efficiency metrics, regardless of whether the source domain VAD training data are available or not. Our results demonstrates the potential of task-irrelevant data as a promising direction for addressing the xVAD problem. {As part of our future work, we will extend our method to enhance it's ability in directly localizing the anomaly in the videos. Another important aspect of VAD models to explore is sensitivity to input manipulations \cite{mumcu2022adversarial, aich2022gama, aich2022leveraging, li2021adversarial} in order to test their robustness.}\\[1.25em] 
\small{
\noindent\textbf{Acknowledgement}. AA and KP were supported by Mitsubishi Electric Research Laboratories. AA and ARC were partially supported by ONR grants N00014-19-1-2264 and N00014-18-1-2252.
}

\onecolumn

% -------- Supplementary Material -------- %
\captionsetup[figure]{list=yes}
\captionsetup[table]{list=yes}
\newpage
\begin{center}
  \vspace*{2\baselineskip}
  \Large\bf{Supplementary material for ``Cross-Domain Video Anomaly Detection\\ without Target Domain Adaptation''}  
\end{center}
\vspace{6\baselineskip}
{
\hypersetup{
    linkcolor=black
}
{\centering
 \begin{minipage}{\textwidth}
     \let\mtcontentsname\contentsname
     \renewcommand\contentsname{\MakeUppercase\mtcontentsname}
     \renewcommand*{\cftsecdotsep}{4.5}
     \noindent
     \rule{\textwidth}{1.4pt}\\[-0.75em]
     \noindent
     \rule{\textwidth}{0.4pt}
     \tableofcontents
     \rule{\textwidth}{0.4pt}\\[-0.70em]
     \noindent
     \rule{\textwidth}{1.4pt}
     \setlength{\cftfigindent}{0pt}
     \setlength{\cfttabindent}{0pt}
     \listoftables
     \listoffigures
 \end{minipage}\par}
}
\clearpage
\renewcommand\thesection{\Alph{section}}
\setcounter{section}{0}
\setcounter{figure}{0}
\setcounter{table}{0}
\resumetocwriting

%---------------- ADDITIONAL DISCUSSION -----% 
\section{Additional Details for {\normalfont{\zvad}}}
\vspace{0.8em}
\paragraph{Additional Implementation Details.}

\begin{wraptable}[8]{R}{8.75cm}
\vspace*{-\baselineskip}
\caption[{Kornia} \cite{riba2020kornia} {augmentation parameters}.]{\textbf{Augmentation parameters.} \ttblue{K} denotes \ttblue{kornia.augmentation}.}
\resizebox{0.5\columnwidth}{!}{%
\begin{tabular}{lcl}
\toprule
\textbf{Operation} &  & \textbf{Kornia Parameters} \\ \cmidrule(r){1-1} \cmidrule(l){3-3} 
\ttblue{K.ColorJitter} &  &  $0.1, 0.1, 0.1, 0.1$ \\
\ttblue{K.RandomAffine} &  & $degrees=360$\\
\ttblue{K.RandomPerspective} &  & $distortion\_scale=0.2$\\ 
\bottomrule
\end{tabular}%
}
\label{tab:kornia_parameters}
\end{wraptable}
We implement \zvad in PyTorch \cite{paszke2019pytorch}. We resize the input frames to $256 \times 256$ and normalize them to the range of $[-1, 1]$. The generator, the discriminator, and the normalcy classifier are trained with the learning rates of $0.0002$, $0.00002$, and $0.00002$, respectively with the Adam \cite{kingma2014adam} optimizer ($\beta_1=0.5, \beta_2=0.999$), following \cite{liu2018future}. The generator takes 4 frames as input and outputs one frame. We drop the last sigmoid layer of $\mathcal{N}(\cdot)$ as suggested in \cite{jolicoeur2018relativistic}. We extracted the frames of all TI datasets at 30 frames/sec. The batch size is set as 8. The training iterations for both SHT and UCFC are set as 5000 in all settings of combinations with TI datasets. Unless otherwise specified, we use the default PyTorch parameters. The average training time is $\sim$2 hours for VAD datasets and $\sim$24 hours for the experiments involving TI datasets on the Nvidia Titan Xp GPUs.\\

\paragraph{Augmentation Parameters.}
Our relative attention affirmation loss $\mathcal{L}_{\text{RAA}}$ requires augmentation of normal frames $\bm{v}$ using Kornia \cite{riba2020kornia} to create augmented normal frames $g(\bm{v})$. We use \ttblue{kornia.augmentation.AugmentationSequential} to apply these augmentation operations sequentially whose 
parameters are listed in Tab. \ref{tab:kornia_parameters}. \ttblue{kornia.augmentation.ColorJitter} has four parameter values that represent factors of ``\textit{brightness}," ``\textit{contrast}," ``\textit{saturation}," and ``\textit{hue}." All the operations have probability parameter $p=1.0$.\\

\paragraph{Location and Size Parameters in Pseudo-Anomaly Synthesis Module.}
Our untrained CNN based Pseudo-anomaly synthesis module $\modO$ creates pseudo-anomalies $\tilde{\bm{v}}$ by pasting cropped object $\bm{M}_{\bm{x}}$ at random location $r_z$ with random size $r_x \times r_y$. We start by initializing a temporary tensor $\overline{\bm{v}}$ with $\bm{v}$. The random location $r_z$ is a rectangular box with coordinates $(b_1, b_2, b_3, b_4)$ \cite{yun2019cutmix}. These are computed as $b_1 = b_x - \nicefrac{b_w}{2}, b_2 = b_x + \nicefrac{b_w}{2}, b_3 = b_y - \nicefrac{b_h}{2}$, and $b_4 = b_y + \nicefrac{b_h}{2}$, where $(b_x, b_y, b_w, b_h)$ are uniformly sampled as follows. If $H$ and $W$ are height and width of $\bm{v}$ respectively, then  $b_x \sim \text{Unif}\big{(}0,W\big{)}, b_y \sim \text{Unif}\big{(}0,H\big{)}, b_w = W\sqrt{1-\beta}, b_h = H\sqrt{1-\beta}$. Here, $b_2>b_1$ and $b_4>b_3$. We then resize $\bm{M}_{\bm{x}}$ and $\bm{M}$ to size $(b_2-b_1)\times (b_4-b_3)$. Finally, only the pixels corresponding to regions where $\bm{M}_{(i,j)}=1$ are replaced in $\overline{\bm{v}}$ to create anomaly frame $\tilde{\bm{v}}$. To handle boundary conditions where $0\leq b_x,b_w\leq W$ and $0\leq b_x,b_w\leq H$, we clip the values to be in the range of $[0, W]$ and $[0, H]$, respectively. Here, $\beta \sim \text{Unif}\big{(}0,1\big{)}$.

\tblue{\paragraph{Evaluation criteria.} For anomaly scores, we follow \cite{liu2018future, lv2021learning} and compute Peak Signal to Noise Ratio (PSNR) \cite{mathieu2015deep} scores per frame and normalize PSNR of all frames in each testing video to the range [0, 1] in order to compare with ground-truth binary labels. Note that we observed such normalization practice (adopted from \cite{liu2018future}) impacts anomaly scores.}

\section{Additional Results on {\normalfont{\zvad}}}
\vspace{0.8em}
\paragraph{Impact of the amount of TI Data.}
We analyzed the impact of the amount of TI data on our \zvad framework in extreme settings. Particularly, we evaluated \zvad when the amount of videos of TI datasets (HMDB and UCF101) is close to the number of training videos available in the VAD datasets. With 0.5\%, 1\%, 2\%, 4\%, and 8\% of HMDB data, we observed an average cross-domain AUC performance of 74.99\% on Ped1, 93.82\% on Ped2, and 79.49\% on Ave. A similar observation was made on UCF101 (0.0625\%, 0.125\%, 0.315\%, 0.63\%, and 1.25\% of data resulted in average cross-domain AUC performance of 74.61\% on Ped1, 94.17\% on Ped2, and 79.46\% on Ave). This demonstrates that almost SOTA cross-domain performance on the current VAD datasets is achievable even with an extremely low amount of TI data.\\

\paragraph{Relevancy among VAD data.} We followed \cite{peng2018zero,fu2015zero} for the relevancy analysis between the TI to target domain (Ave, Ped1/2) VAD data. We observed higher relevancy scores among SHT (to Ave: 0.241, to Ped1/2: 0.250) and UCFC (to Ave: 0.201, to Ped1/2: 0.167) compared to average TI (to Ave: 0.186, to Ped1/2: 0.138). This confirms: TI data is indeed less relevant to VAD data. \\

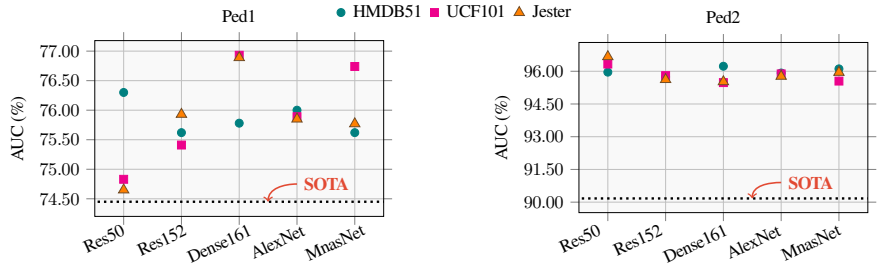
\begin{figure}[!ht]
    \begin{minipage}[c]{0.75\columnwidth}
    \centering
    \vspace{-0.5\baselineskip}
    \begin{subfigure}[t]{0.445\textwidth}
        \raggedright
        \pgfplotsset{every x tick label/.append style={font=\small, rotate=25, xshift=-1.5ex, yshift=2ex}}

\pgfplotsset{y tick label style={/pgf/number format/fixed zerofill, /pgf/number format/precision=2, font=\small}}

\pgfplotstableread[col sep=&, header=true]{
  network      & hmdb   & ucf    & jester
  Res50     & 76.30  & 74.83  & 74.65
  Res152    & 75.62  & 75.41  & 75.93
  Dense161  & 75.78  & 76.93  & 76.89
  AlexNet      & 76.00  & 75.89  & 75.85
  MnasNet      & 75.62  & 76.74  & 75.77
}\ped % ped1

\begin{tikzpicture}[scale=0.75, transform shape, every node/.append style={font=\small}]
    \begin{axis}
      [
        xtick=data,
        xticklabels from table={\ped}{network},  
        width= 1.15\textwidth,
        height = 47mm,
        nodes near coords align={vertical},
        xmin=0.5, xmax=5.5,
        ytick={74,74.5,...,77.5},
        ylabel={AUC (\%)},
        title={Ped1},
        legend style={at={(1.25,1.25)}, anchor=north, column sep=0.5ex, draw=none, only marks},
        legend columns=-1,
        axis background/.style={fill=gray!5},
        grid=both,
        ytick align=outside,
        xtick align=outside
      ]
      \addplot+[color=teal, mark options={fill=teal}] table [y=hmdb, x expr=\coordindex+1, only marks] {\ped};
      \addlegendentry{HMDB51}
      \addplot+[color=magenta, mark options={fill=magenta}] table [y=ucf, x expr=\coordindex+1, only marks] {\ped};
      \addlegendentry{UCF101}
      \addplot+[mark=triangle*, mark size=3pt, mark options={fill=orange}, only marks] table [y=jester, x expr=\coordindex+1, only marks] {\ped};
      \addlegendentry{Jester}
      \addplot[mark=none, very thick, dotted, black, domain=0:6] {74.45};
      \draw [-latex, MyStyleSmooth, color=ped1-color] (4.0,74.75) node[right]{\textbf{SOTA}} to[out=-180,in=90] (3.5,74.45);
    \end{axis}
  \end{tikzpicture}
        \label{fig:random_nw_ped1}
    \end{subfigure}%
    ~ \hspace{1em}
    \begin{subfigure}[t]{0.445\textwidth}
        \raggedright
        \pgfplotsset{every x tick label/.append style={font=\small, rotate=25, xshift=-2.5ex, yshift=2ex}}

\pgfplotsset{y tick label style={/pgf/number format/fixed zerofill, /pgf/number format/precision=2, font=\small}}

\pgfplotstableread[col sep=&, header=true]{
  network      & hmdb   & ucf    & jester
  Res50     & 95.96  & 96.33  & 96.67
  Res152    & 95.74  & 95.80  & 95.62
  Dense161  & 96.23  & 95.48  & 95.52
  AlexNet      & 95.92  & 95.88  & 95.77
  MnasNet      & 96.11  & 95.55  & 95.95
}\ped % ped2

  \begin{tikzpicture}[scale=0.75, transform shape, every node/.append style={font=\small}]
    \begin{axis}
      [
        xtick=data,
        xticklabels from table={\ped}{network},      
        width= 1.15\textwidth,
        height = 46mm,
        nodes near coords align={vertical},
        xmin=0.5, xmax=5.5,
        ylabel={AUC (\%)},
        ytick={90,91.5,93,94.5,96,97.5},
        title={Ped2},
        legend style={at={(0.5,1.25)}, anchor=north, column sep=1ex, draw=none, only marks},
        legend columns=-1,
        axis background/.style={fill=gray!5},
        grid=both,
        ytick align=outside,
        xtick align=outside
      ]
      \addplot+[color=teal, mark options={fill=teal}] table [y=hmdb, x expr=\coordindex+1, only marks] {\ped};
      %\addlegendentry{HMDB51}
      \addplot+[color=magenta, mark options={fill=magenta}] table [y=ucf, x expr=\coordindex+1, only marks] {\ped};
      %\addlegendentry{UCF101}
      \addplot+[mark=triangle*, mark size=3pt, mark options={fill=orange}, only marks] table [y=jester, x expr=\coordindex+1, only marks] {\ped};
      %\addlegendentry{Jester}
      \addplot[mark=none, very thick, dotted, black, domain=0:6] {90.17};
      \draw [-latex, MyStyleSmooth, color=ped1-color] (4.0,90.90) node[right]{\textbf{SOTA}} to[out=-180,in=90] (3.5,90.17);
    \end{axis}
\end{tikzpicture}
        \label{fig:random_nw_ped2}
    \end{subfigure}%
    \end{minipage}%
    \begin{minipage}[c]{0.25\columnwidth}
    \vspace{-1.5\baselineskip}
    \caption[Impact of $\mathcal{R}(\cdot)$ in \zvad (Ped1 and Ped2).]{\textbf{Impact of $\mathcal{R}(\cdot)$} in \zvad. The source is SHT.}
    \label{fig:supp_random_nw}
    \end{minipage}%
    %\vspace{-1.5\baselineskip}
\end{figure}

\paragraph{More results on the impact of randomly initialized networks for Pseudo-Anomaly Synthesis.}
We analyzed the impact of the randomly initialized network $\mathcal{R}(\cdot)$ on our untrained CNN based pseudo-anomaly synthesis module. In Fig. \ref{fig:supp_random_nw}, it can be observed that our \zvad method outperforms the state-of-the-art (SOTA) xVAD works on the Ped1 and Ped2 datasets in the zero-shot settings when the source is SHT irrespective of kind of randomly initialized network $\mathcal{R}(\cdot)$ employed to extract objects from all our TI datasets.

\begin{table}[ht]
\begin{minipage}[t]{0.7\columnwidth}
\paragraph{More results on same-dataset testing.} We beat our baselines in the same-dataset testing in all VAD and TI combination scenarios as shown in Tab. \ref{tab:supp_same_dataset_exp}. \tblue{We also compare with more state-of-the-art unsupervised VAD methods under the same-dataset setting in Tab. \ref{tab:supp_same_data}.} \\
\caption[Same-dataset testing on the SHT$_{\text{dc}}$]{\textbf{Same-dataset testing on the SHT$_{\text{dc}}$ dataset.} We beat our baselines in all the source domain data settings.}
\label{tab:supp_same_dataset_exp}
\resizebox{\columnwidth}{!}{%
\begin{tabular}{cccc}
\toprule
\textbf{VAD Training data} & \textbf{Input to $\modO$} & \textbf{Method} &\textbf{AUC} (\%) \textbf{on SHT}$_\text{dc}$ \\
\midrule
SHT$_{\text{dc}}$ &N/A &\rgan (paper) &70.11 \\
SHT$_{\text{dc}}$ &N/A &\mpn (code) &67.47 \\
\rowcolor{red!10}
SHT$_{\text{dc}}$ &SHT$_{\text{dc}}$ &\zvado &70.73 \\
\rowcolor{red!10}
SHT$_{\text{dc}}$ &HMDB &\zvado &\textbf{70.85} \\
\rowcolor{red!10}
SHT$_{\text{dc}}$ &UCF101 &\zvado &70.80 \\
\rowcolor{red!10}
SHT$_{\text{dc}}$ &Jester &\zvado &70.50 \\
\bottomrule
\end{tabular}%
}
\end{minipage}%
\hfill
\begin{minipage}[t]{0.25\columnwidth}
\vspace{-0.8\baselineskip}
\caption[Additional same dataset testing comparison]{\textbf{Additional same dataset testing comparison.} The best and second best AUC are marked in \textbf{bold} and \underline{underline}, respectively.}
\tiny{
\resizebox{\columnwidth}{!}{
\begin{tabular}{cccc}
	\toprule
	\textbf{Methods} & \textbf{Ped2} & \textbf{Ave} & \textbf{SHT} \\
	\midrule
    MPPCA~\cite{kim2009observe} & 69.3 & - & - \\
    MPPC+SFA~\cite{kim2009observe} & 61.3 & - & - \\
    MDT~\cite{mahadevan2010anomaly} & 82.9 & - & - \\
    ConvAE~\cite{hasan2016learning} & 85.0 & 80.0 & 60.9 \\
    TSC~\cite{luo2017revisit} & 91.0 & 80.6 & 67.9 \\
    StackRNN~\cite{luo2017revisit} & 92.2 & 81.7 & 68.0 \\
    MT-FRCN~\cite{hinami2017joint} & 92.2 & - & - \\
    Unmasking~\cite{tudor2017unmasking} & 82.2 & 80.6 & - \\
    Frame-Pred~\cite{liu2018future} & 95.4 & 85.1 & 72.8 \\
    AMC~\cite{nguyen2019anomaly} & 96.2 & 86.9 & -\\
    MemAE~\cite{gong2019memorizing} & 94.1 & 83.3 & 71.2 \\	
    SDOR~\cite{pang2020self} & 83.2 & - & -\\
    \rgan & 96.2 & 85.8 & \textbf{77.9} \\
    LMN~\cite{park2020learning} & \textbf{97.0}  & \ul{88.5}&  70.5\\
    \mpn & {96.9}  &  \textbf{89.5} &  \ul{73.8}\\
    \rowcolor{red!10}
    \zvad & \ul{96.95} & 83.8 & 71.6\\
	\bottomrule
\end{tabular}}}
\label{tab:supp_same_data}
\end{minipage}
\end{table}

\begin{wraptable}[5]{r}{2.75cm}
\vspace{-\baselineskip}
\hspace{-1.5em}
\centering
\resizebox{0.175\textwidth}{!}{
    \begin{tabular}{rc} 
    \hline
    \textbf{Ratios} & \textbf{SHT}$_{dc}$\\
    \hline
     \mpn &67.47\\
    \hline
     $(\alpha_{\text{n}}, \alpha_{\text{rn}}) = (1,0.01)$ & \textbf{70.85}\\
     $(\alpha_{\text{n}}, \alpha_{\text{rn}}) =  (1,0.1)$ & 69.49 \\
     $(\alpha_{\text{n}}, \alpha_{\text{rn}}) =  (0.1,0.1)$ & 69.95 \\
     $(\alpha_{\text{n}}, \alpha_{\text{rn}}) =  (0.01,0.01)$ & 70.37 \\
     \hline
    \end{tabular}}
\end{wraptable} 
\vspace{\baselineskip}
\paragraph{Ablation analysis.} \zvad is not too sensitive to the loss ratios and Table (on \textit{right}) validates this point. For our backbone GAN, we use exact same ratios as suggested in \cite{liu2018future}. For the proposed normalcy classifier, we \textit{do not} use ratios for our losses $\mathcal{L}_{\text{AA}}$ and $\mathcal{L}_{\text{RAA}}$ (\ie set as 1). Finally, the effect of ratios $\alpha_{\text{n}}$ on $\mathcal{L}_{\text{N}}$ and $\alpha_{\text{rn}}$ on $\mathcal{L}_{\text{RN}}$ is shown. All cases show better AUC than SOTA \mpn.

%---------------- DATASETS ------------------% 
\section{Examples from Datasets}
\begin{figure}[ht]
    \centering
    \begin{subfigure}[t]{\textwidth}
        \centering
        \includegraphics[width=\textwidth]{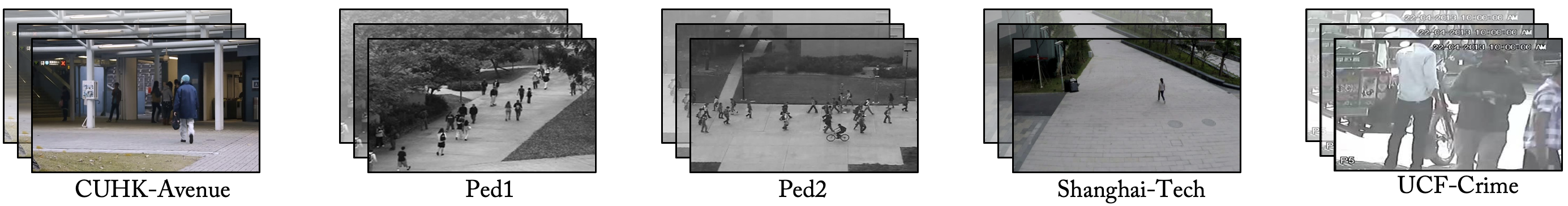}
        \caption{VAD datasets}
        \label{fig:vad_data}
    \end{subfigure}%
    \\[0.5em]
    \begin{subfigure}[t]{\textwidth}
        \centering
        \includegraphics[width=0.6\textwidth]{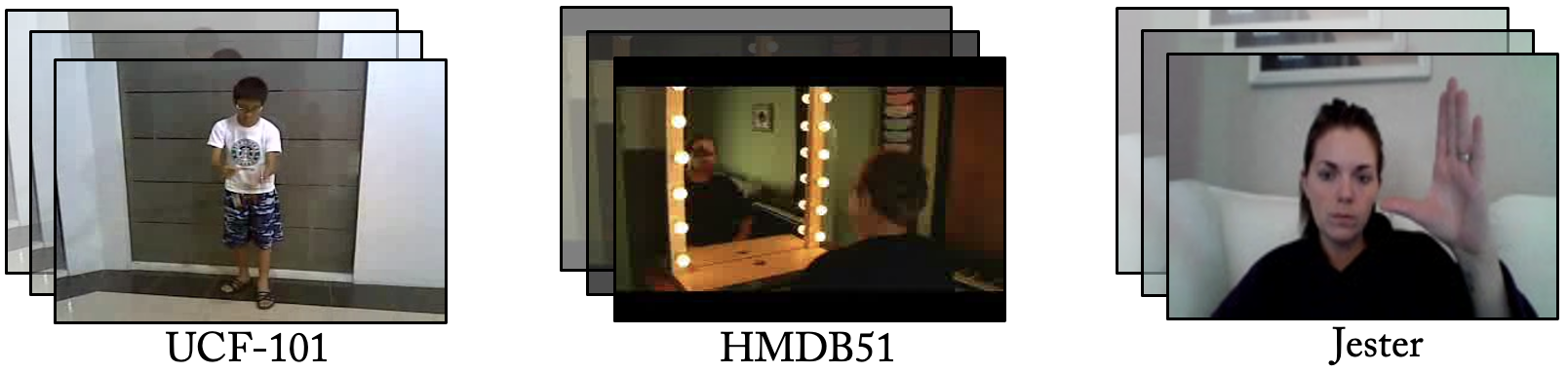}
        \caption{TI datasets}
        \label{fig:ti_data}
    \end{subfigure}
    \caption[Examples from datasets used in our paper.]{\textbf{Examples from VAD and TI datasets.} We visualize some examples of videos used for experiments in our paper.}
    %\vspace{-2\baselineskip}
\end{figure}

We provide some video examples of the VAD datasets (SHT, UCFC, Ped1, Ped2, and Ave in Fig.\ref{fig:vad_data}) and TI datasets (HMDB, UCF101, and Jester in Fig.\ref{fig:ti_data}) listed in \texttt{Tab.2} of the main manuscript. 

%---------------- QUALITATIVE RESULTS -------% 
\section{More Qualitative Results}
\begin{figure}[ht]
    \centering \includegraphics[width=0.975\textwidth]{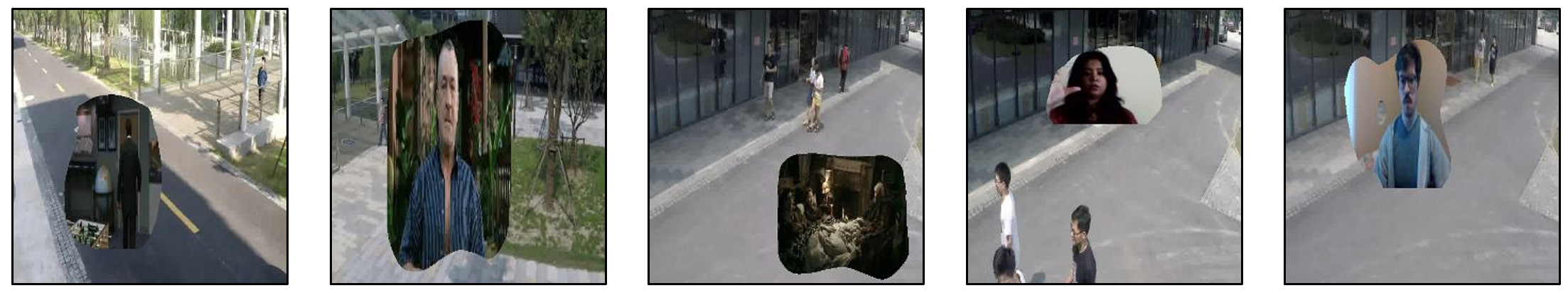}
    \caption[More examples of pseudo-abnormal frames.]{\textbf{Pseudo-abnormal frames.} We present examples of pseudo-abnormal frames generated using our proposed untrained CNN based pseudo-anomaly synthesis module.}
    \label{fig:supp_pseudo}
\end{figure}

\begin{figure}[!ht]
    \centering \includegraphics[width=0.975\textwidth]{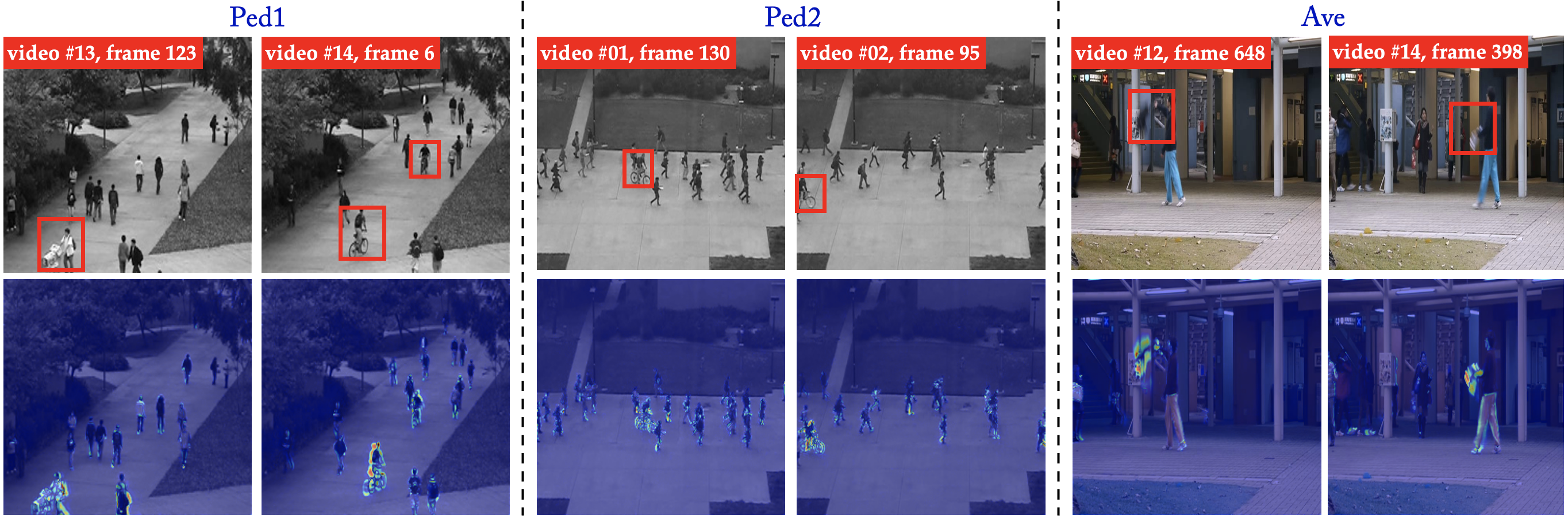}
    \caption[More examples of difference maps indicating anomalies.]{\textbf{Difference maps.} We show more examples of difference maps obtained from \zvad (source: SHT). The lighter colors in difference map mean larger prediction error indicating anomalies. Red boxes indicate ground truth anomalies. Best viewed in color.}
    \label{fig:supp_error_maps}
\end{figure}

We show additional examples of pseudo-abnormal frames created using our pseudo-anomaly module in Fig. \ref{fig:supp_pseudo} and difference maps from three different datasets indicating anomalies in Fig. \ref{fig:supp_error_maps}.

%%%%%%%%% REFERENCES
\twocolumn
\FloatBarrier

{\small
\bibliographystyle{unsrt}
\bibliography{egbib}
}

\end{document}